\documentclass[sigconf,10pt,nonacm]{acmart}
\AtBeginDocument{%
  \fancypagestyle{firstpagestyle}{%
    \fancyhf{}
    \fancyhead[C]{\Small
      PREPRINT: Accepted to The Tenth ACM/IEEE Symposium on Edge Computing (SEC'25).\quad
      Final version: \url{https://doi.org/10.1145/3769102.3770614}}
    
  }%
}

\usepackage{hyperref}
\usepackage{url}
\usepackage{adjustbox}
\usepackage{multirow}
\usepackage{subfigure}
\usepackage{threeparttable}
\usepackage{siunitx}
\usepackage{pifont}
\newcommand{\xmark}{\textcolor{red}{\ding{55}}}
\usepackage{xcolor}

\usepackage{enumitem}
\setlist[itemize]{leftmargin=*}
\setlist[enumerate]{leftmargin=*}

\usepackage{listings}
\lstdefinestyle{cppstyle}{
  language=C++,
  basicstyle=\ttfamily\normalsize,
  keywordstyle=\color{blue},
  commentstyle=\color{gray},
  stringstyle=\color{red},
  numbersep=5pt,
  breaklines=true,
  frame=single,
  columns=fullflexible,
  keepspaces=true,
  showstringspaces=false
}

\usepackage{mdframed}
\newmdenv[
  backgroundcolor=green!6,
  linewidth=0.4pt, 
  innertopmargin=3pt, 
  innerbottommargin=3pt, 
  skipabove=4pt, 
  skipbelow=4pt, 
  innerleftmargin=3pt,
  innerrightmargin=3pt
]{graybox}

\begin{document}

\title{\textsc{lm-Meter}: Unveiling Runtime Inference Latency for On-Device Language Models}

\author{Haoxin Wang}
\email{haoxinwang@gsu.edu}
\affiliation{%
  \institution{Georgia State University}
  \city{Atlanta}
  \state{GA}
  \country{USA}
}

\author{Xiaolong Tu}
\email{xtu1@student.gsu.edu}
\affiliation{%
  \institution{Georgia State University}
  \city{Atlanta}
  \state{GA}
  \country{USA}}

\author{Hongyu Ke}
\email{hke3@student.gsu.edu}
\affiliation{%
  \institution{Georgia State University}
  \city{Atlanta}
  \state{GA}
  \country{USA}
}

\author{Huirong Chai}
\email{hchai4@student.gsu.edu}
\affiliation{%
 \institution{Georgia State University}
  \city{Atlanta}
  \state{GA}
  \country{USA}}

\author{Dawei Chen}
\email{dawei.chen1@toyota.com}
\affiliation{%
 \institution{Toyota InfoTech Labs}
  \city{Mountain View}
  \state{CA}
  \country{USA}}

\author{Kyungtae Han}
\email{kt.han@toyota.com}
\affiliation{%
  \institution{Toyota InfoTech Labs}
  \city{Mountain View}
  \state{CA}
  \country{USA}}

\renewcommand{\shortauthors}{Haoxin Wang et al.}

\begin{abstract}
{\sloppy
Large Language Models (LLMs) are increasingly integrated into everyday applications, but their prevalent cloud-based deployment raises growing concerns around data privacy and long-term sustainability. 
Running LLMs locally on mobile and edge devices (on-device LLMs) offers the promise of enhanced privacy, reliability, and reduced communication costs. 
However, realizing this vision remains challenging due to substantial memory and compute demands, as well as limited visibility into performance-efficiency trade-offs on resource-constrained hardware.
We propose \textsc{lm-Meter}, the first lightweight, online latency profiler tailored for on-device LLM inference. \textsc{lm-Meter} captures fine-grained, real-time latency at both phase (e.g., embedding, prefill, decode, softmax, sampling) and kernel levels without auxiliary devices.
We implement \textsc{lm-Meter} on commercial mobile platforms and demonstrate its high profiling accuracy with minimal system overhead, e.g., only 2.58\% throughput reduction in prefill and 0.99\% in decode under the most constrained \texttt{Powersave} governor.
Leveraging \textsc{lm-Meter}, we conduct comprehensive empirical studies revealing phase- and kernel-level bottlenecks in on-device LLM inference, 
quantifying accuracy-efficiency trade-offs, and identifying systematic optimization opportunities. 
\textsc{lm-Meter} provides unprecedented visibility into the runtime behavior of LLMs on constrained platforms, laying the foundation for informed optimization and accelerating the democratization of on-device LLM systems.
Code and tutorials are available at \href{https://github.com/amai-gsu/LM-Meter}{\color{magenta}\texttt{github.com/amai-gsu/LM-Meter}}.
\par}
\end{abstract}

\keywords{Large Language Models, On-Device AI, Edge Computing}

\maketitle

\section{Introduction}
Large language models (LLMs), such as GPT-series~\cite{brown2020language, achiam2023gpt}, LLaMA~\cite{touvron2023llama}, and DeepSeek~\cite{guo2025deepseek,liu2024deepseek}, have recently garnered significant attention for their impressive capabilities in natural language understanding, generation, and reasoning across a broad spectrum of tasks. By scaling to billions or even trillions of parameters, these models have achieved state-of-the-art performance in applications including machine translation, dialogue systems, and code generation~\cite{devlin2019bert, radford2019language, chen2021evaluating, chowdhery2023palm}.
Today, the predominant approach to deploying LLMs is cloud-based, relying on centralized inference services hosted in large-scale data centers. These services are typically powered by specialized hardware accelerators, custom software stacks, and a suite of system-level optimizations, such as tensor parallelism~\cite{shoeybi2019megatron}, speculative decoding~\cite{chen2023accelerating}, and memory-efficient KV cache management~\cite{dao2022flashattention}, to support emerging use cases with high-throughput, low-latency inference at scale.

Despite the benefits of deploying LLMs in cloud environments, this paradigm has pushed the computational boundaries of cloud infrastructure and exacerbates concerns over its long-term sustainability (\S\ref{ssc:why}). Moreover, issues surrounding data privacy and user data custody have become increasingly prominent. For instance, many private enterprises and government agencies restrict the use of cloud-hosted LLM services for processing sensitive or proprietary information due to potential risks involving data leakage, regulatory non-compliance, and lack of control over data residency.
These concerns underscore a growing demand for local LLM accessibility on mobile and edge devices (i.e., on-device LLMs), such as smartphones and Internet-of-Things (IoT) devices, to enable broader and more privacy-preserving integration of LLMs into users’ everyday lives.
On-device LLMs offer several compelling benefits, including enhanced data privacy, greater reliability under intermittent connectivity, and reduced communication costs. They also enable user-centric personalization and context-aware adaptation, unlocking new opportunities for responsive, interactive applications directly on personal consumer devices \cite{laskaridis2024future}.

However, deploying LLMs effectively and efficiently on mobile and edge hardware remains largely under-explored. First, the parameter sizes of most state-of-the-art LLMs scale into the billions, demanding substantial memory capacity to accommodate model weights, intermediate activations, and inference computations, often exceeding the limited compute and memory resources available on mobile and edge devices.
Second, most existing model architectures and software frameworks are designed and optimized for cloud or high-performance computing environments. 
Third, there is a limited understanding of the fundamental trade-offs between performance and efficiency in on-device LLM inference. Critical bottlenecks, such as task-dependent throughput limitations and latency in specific inference phases and kernels, remain insufficiently characterized.
Existing literature provides minimal insight into opportunities for systematic optimization, adaptive runtime strategies, kernel-level improvements, or efficient scheduling under tight resource budgets.
More importantly, the lack of lightweight, real-time profiling tools further exacerbates these challenges, restricting the ability to comprehensively analyze and optimize on-device LLM inference across algorithmic, kernel, and hardware levels.

To this end, we propose \textsc{lm-Meter}, the first lightweight online runtime latency profiler for on-device LLM inference. \textsc{lm-Meter} enables accurate, real-time latency measurement at both phase (e.g., embedding, prefill, decode, softmax, and sampling) and kernel granularity on consumer-grade mobile and edge platforms, without auxiliary devices. 
We envision \textsc{lm-Meter} as a foundation for studying runtime bottlenecks and deployment constraints, addressing a key barrier to the practical democratization of on-device LLMs. Our main contributions are summarized below:
\begin{itemize}
\sloppy
\item We design and implement \textsc{lm-Meter}, a lightweight online runtime latency profiler tailored for on-device LLM inference. Through evaluations on multiple consumer-grade mobile devices, we demonstrate that \textsc{lm-Meter} accurately captures both phase-level and kernel-level latencies with minimal system overhead, validating its practicality for real-world deployment.
\par

\item Leveraging \textsc{lm-Meter}, we conduct a phase-oriented empirical study to systematically analyze the Pareto frontier of performance-efficiency trade-offs across diverse benchmarking tasks, popular open-source LLM suites, varying model sizes, and mobile platforms. 
We further conduct a systematic evaluation to characterize how quantization shifts these trade-offs by proposing a new metric, the Harmonic Quantization score ($\mathcal{HQ}$).

\item We perform a kernel-oriented empirical study to uncover fine-grained execution bottlenecks and resource utilization patterns at the kernel level. 

\item Based on our empirical analyses, we identify critical bottlenecks in on-device LLM inference and distill actionable insights for both system- and model-level optimizations, informing the design of more efficient, scalable, and responsive on-device LLM systems.
\end{itemize}

\section{Background and Motivation}
\label{sc:motv}

\subsection{Why Democratizing On-Device LLMs}
\label{ssc:why}
\textbf{Sustainability.} 
Prominent implementations of contemporary LLM products, such as ChatGPT \cite{chatgpt} and Google NotebookLM \cite{notebooklm}, primarily rely on cloud-based infrastructures due to their massive model sizes (ranging from billions to trillions of parameters) and substantial computational demands. As human reliance on LLMs continues to grow, envisioning a future where it becomes deeply integrated into daily life, supporting both front-end interactions, such as personalized companionship \cite{wang2022leaf,wang2020user,wang2017v,ke2024carboncp}, and back-end applications like recommendation systems, its usage could constitute $\sim5\%$ of an individual’s daily time \cite{liu2024mobilellm}. 
In such a scenario, operating a \textit{single} LLM model (e.g., GPT-4) at a processing rate of 50 tokens per second to serve the global population would require the deployment of roughly 100 million H100 GPUs (each capable of 60 trillion floating-point operations per second) \cite{H100}. 
This immense computational scale required, excluding network overhead, would be equivalent to the infrastructure of approximately \textit{160 Meta-scale} companies \cite{MetaGPUs}.

\textbf{User privacy.} Centralized LLM services require user data, often highly sensitive, to traverse public networks and be processed in remote data centers, creating significant attack surfaces. Privacy risks are most acute in domains that handle regulated or confidential information, including healthcare (e.g., patient health data), finance, and enterprise productivity workflows containing proprietary IP. On-device LLMs eliminate this exposure by keeping both the model and the data local, dramatically reducing the attack surface and aligning with regulations such as HIPAA \cite{HIPAA}. This privacy-by-design paradigm makes local deployment not merely a technical optimization but a prerequisite for user trust. Removing the dependency on continuous connectivity also improves reliability in bandwidth‑constrained or offline scenarios.

In summary, transitioning LLM computations to mobile and edge devices, including smartphones, personal computers, and IoT devices, is essential for achieving improved sustainability, enhanced privacy, and robust network independence by bringing computation closer to end users.

\subsection{Why a Lightweight Online Profiler Matters}
\label{ssc:needs}

Despite its promising potential, on-device LLM deployment remains largely under-explored and poses formidable research challenges. Bridging the gap between their immense computational demands and the resource constraints of mobile and edge devices requires fundamental breakthroughs. These constraints include restricted memory and compute capacity, limited energy reserves (e.g., battery life or intermittent power), and insufficient support for parallelism \cite{10876858}.

{\sloppy
We conduct a pilot benchmark to validate the above concerns using one of the most popular open-source LLM suites, \texttt{Qwen}~\cite{qwen}. 
Three generations, \texttt{Qwen-1.5}, \texttt{Qwen-2}, and \texttt{Qwen-2.5}, are evaluated in their smallest 0.5B parameter configurations to ensure they can run unmodified (no quantization or fine-tuning)
on contemporary mobile hardware. 
Table \ref{tb:mot1} presents accuracy on three benchmark tasks (ARC-Challenge, HellaSwag, GSM8K) and token-generation throughput measured on a Google Pixel 8 Pro.
Interestingly, the results reveal a widening accuracy-efficiency gap. While each successive generation improves accuracy (e.g., $+4.4$ percentage point (pp) on HellaSwag and $+24.5$ pp on GSM8K from \texttt{Qwen-1.5} to \texttt{Qwen-2.5}), on-device decode throughput drops by 24\%, from 22.76 tokens/s to 17.29 tokens/s. This trend underscores a prevailing research emphasis on accuracy gains, often at the expense of runtime efficiency, and directly conflicts with the constraints of on-device LLMs.
\par
}

Consequently, a comprehensive and systematic understanding of on-device LLM bottlenecks is now more urgent than ever. To this end, we advocate for a lightweight, online profiler tailored to on-device LLMs. Unlike traditional offline profilers, an online profiler enables real-time collection of fine-grained performance metrics (e.g., latency, memory usage) during live model execution. These data can be immediately accessed by the application or operating system, enabling dynamic optimizations, adaptive scheduling, and informed system-level decisions, key steps toward the practical democratization of LLMs at the edge.
Below, we highlight two key benefits of a lightweight, online profiler:

\textbf{Empirical study on on-device LLMs.} A lightweight profiler is essential for enabling accurate, empirical analysis of LLM behavior on mobile and edge devices. Unlike cloud-based profiling tools that can tolerate additional system overhead, on-device studies demand minimal interference with the model's runtime. A profiler that introduces significant overhead distorts key performance signals such as latency, memory usage, and energy consumption, leading to misleading conclusions about real-world performance. By keeping overhead low and integrating seamlessly with existing runtime pipelines, a lightweight online profiler allows researchers to faithfully observe the system under typical usage conditions.

\textbf{Online runtime performance prediction.}
LLMs generate output in an autoregressive manner, producing a token at a time during decoding. For each token, the model embeds the input, attends over all previously generated tokens stored in a growing key-value (KV) cache, processes the result through transformer layers, samples the next token, and updates the KV cache. The computational cost of this process increases approximately linearly with sequence length, while hardware-level factors on mobile and edge devices, such as dynamic frequency scaling and thermal throttling, tend to evolve gradually. 
These properties make throughput prediction for on-device LLM both feasible and valuable. A lightweight profiler that measures per-token latency in real time can exploit these trends to predict future throughput over short horizons. Such predictions are critical for enabling dynamic scheduling decisions and system adaptation.

In addition, recent studies have shown that kernel-level prediction mechanisms can achieve state-of-the-art accuracy in predicting both latency \cite{zhang2021nn,feng2024litepred} and energy consumption \cite{tu2023unveiling,mallik2023epam,tu2023deepen2023,tu2025greenauto} for conventional deep learning models. These approaches benefit from the fact that kernels serve as fundamental scheduling units and encapsulate many framework-level optimizations.
On-device LLM systems can adopt similar strategies by accurately profiling fine-grained, kernel-level execution data.

\begin{table}[t]
\centering
\caption{Comparison of accuracy and on-device decode throughput of \texttt{Qwen} model generations.}
\label{tb:mot1}
\begin{adjustbox}{max width=\columnwidth}
\begin{threeparttable}
\begin{tabular}{l|c|c|c|c}
\toprule[1pt]\midrule[0.3pt]
\multirow{2}{*}{Models} & \multicolumn{3}{c|}{Accuracy $\uparrow$} & Throughput \multirow{2}{*}{$\uparrow$}\\ \cline{2-4}
                        &ARC-Challenge&HellaSwag&GSM8K& (tokens/s)\\
\hline
\texttt{Qwen-1.5-0.5B}          &0.293    & 0.491    & 0.171  &\textbf{22.76}\\ 
\texttt{Qwen-2-0.5B}            &0.310    & 0.491    & 0.364 & 17.33\\ 
\texttt{Qwen-2.5-0.5B} &\textbf{0.356}& \textbf{0.522}  & \textbf{0.416} & 17.29\\ 
\midrule[0.3pt]\bottomrule[1pt]
\end{tabular}
\begin{tablenotes}[flushleft]\small
\item[*] Throughput data are collected on Google Pixel 8 Pro. Model accuracy on individual tasks is evaluated using the \textit{lm-evaluation-harness}~\cite{evalharness}.
\end{tablenotes}
\end{threeparttable}
\end{adjustbox}
\end{table}

\subsection{Limitations of Existing Profilers}
\label{ssc:profilers}
As shown in Table \ref{tb:profiler}, most existing profilers for mobile and edge platforms were not designed with LLM inference in mind, and therefore fall short of meeting the requirements outlined in Section~\ref{ssc:needs}. 
Specifically, offline profilers such as Android GPU Inspector (AGI)~\cite{AGI} and Perfetto~\cite{perfetto} collect raw execution traces during runtime but rely on extensive post-processing, often on a host machine, to extract meaningful insights. This delayed analysis pipeline, which involves exporting logs, synchronizing metadata, and using external visualization tools, makes them unsuitable for real-time adaptation and on-device scheduling. Furthermore, these tools lack kernel-level visibility, which is essential for capturing the fine-grained behaviors of LLM execution.

Recently, several online profilers have been developed for on-device learning scenarios, including MELTing Point~\cite{laskaridis2024melting} and nnPerf~\cite{nnPerf}. However, both exhibit several limitations. MELTing Point reports only the average duration of each GPU kernel in isolation, while capturing cumulative runtime, it lacks timeline visibility and does not provide fine-grained start/end timestamps or show how kernels are sequenced and interleaved. As a result, it fails to reveal runtime behaviors such as kernel launch delays, inter-kernel gaps, or scheduling contention. More importantly, enabling MELTing Point’s kernel-level tracing introduces substantial latency overhead (see Fig.~\ref{fig:overhead_exam}), making it unsuitable for accurate kernel-granular analysis of on-device LLM inference.
nnPerf~\cite{nnPerf}, on the other hand, was primarily designed for smaller convolutional or feedforward models, such as MobileNet, EfficientNet, and ResNet. While effective for traditional DNNs, it lacks support for the unique computational characteristics of LLMs, which involves dynamic input lengths, token-level scheduling bottlenecks, and interleaved execution phases, e.g., embedding, prefill, autoregressive decoding, softmax, and token sampling.



\textit{To this end, we propose \textsc{lm-Meter}, a lightweight, online profiler that enables LLM phase-aware and fine-grained kernel-level latency tracing, specifically designed for on-device LLM inference. \textsc{lm-Meter} equips researchers and system designers with transparent and deeper visibility into runtime behavior of LLMs on resource-constrained platforms, facilitating informed optimization, adaptive scheduling, and accelerating the democratization of intelligent on-device systems.}

\begin{table}[t]
\centering
\caption{\textsc{lm-Meter} vs. existing profilers for mobile and edge hardware.}
\label{tb:profiler}
\begin{adjustbox}{max width=\columnwidth}
\begin{threeparttable}
\begin{tabular}{l|c|c|c|c|c|c|c}
\toprule[1pt]\midrule[0.3pt]
\multirow{2}{*}{Profilers} & Support  & Phase     & Kernel- & Timeline       & Low      & Real-time & No-host \\
                           & LLMs     & awareness & level   & visibility & overhead & output    & machine \\
\hline
AGI~\cite{AGI} &{\color{black} \checkmark}&{\color{black} \checkmark}& \xmark  &{\color{black} \checkmark}&{\color{black} \checkmark}& \xmark &  \xmark \\ \hline

Perffeto~\cite{perfetto} &{\color{black} \checkmark}&{\color{black} \checkmark}& \xmark  &{\color{black} \checkmark}&{\color{black} \checkmark}&  \xmark & \xmark  \\ \hline

TFLite & \multirow{2}{*}{\xmark} & \multirow{2}{*}{\xmark}  &\multirow{2}{*}{\color{black} \checkmark}& \multirow{2}{*}{\xmark}  &\multirow{2}{*}{\color{black} \checkmark}& \multirow{2}{*}{\xmark}  & \multirow{2}{*}{\xmark}  \\
benchmark~\cite{tflite} &   &   &   &   &   &   &   \\ \hline
nnPerf~\cite{nnPerf}& \xmark  & \xmark  &{\color{black} \checkmark}&{\color{black} \checkmark}&{\color{black} \checkmark}&{\color{black} \checkmark}&{\color{black} \checkmark}\\ \hline
MELT~\cite{laskaridis2024melting} &{\color{black} \checkmark}&{\color{black} \checkmark}&{\color{black} \checkmark} & \xmark  &  \xmark &{\color{black} \checkmark}&{\color{black} \checkmark}\\
\hline
\textbf{\textsc{lm-Meter}} & \multirow{2}{*}{\color{black} \checkmark} &\multirow{2}{*}{\color{black} \checkmark}&\multirow{2}{*}{\color{black} \checkmark}&\multirow{2}{*}{\color{black} \checkmark}&\multirow{2}{*}{\color{black} \checkmark}&\multirow{2}{*}{\color{black} \checkmark}&\multirow{2}{*}{\color{black} \checkmark} \\
 \textbf{(Ours)}   &   &   &   &   &   &   &   \\
 
\midrule[0.3pt]\bottomrule[1pt]
\end{tabular}
\begin{tablenotes}[flushleft]\small
\item[*] Timeline visibility: whether the profiler can log and reconstruct an execution timeline with start/end timestamps for each kernel, data transfer, and GPU idle period.
\item[$\dagger$] Phase awareness: whether the profiler can distinguish major LLM inference phases and report per-phase latency attribution and bottleneck analysis.
\item[$\bigstar$] No-host machine: whether the profiler operates independently on-device without relying on an auxiliary machine to run or process measurements.
\end{tablenotes}
\end{threeparttable}
\end{adjustbox}
\end{table}

\section{\textsc{lm-Meter} Design}
\label{sc:design}


\subsection{Phase-level Latency Profiling}
 
Phase-level inference latency refers to the runtime spent in distinct functional stages of LLM execution, such as embedding, prefill, decode, softmax, and token sampling. 
Unlike end-to-end (E2E) latency, which offers a coarse aggregate metric, phase-level profiling provides a semantically meaningful decomposition of the inference pipeline. 
This finer granularity brings several advantages. First, it enables precise bottleneck diagnosis by identifying which stages dominate latency, thereby guiding targeted optimizations. Second, it uncovers how different hardware architectures interact with individual phases, supporting platform-aware deployment and tuning. Third, it prevents misleading conclusions that may arise from aggregated E2E metrics by revealing heterogeneous performance characteristics across phases. Lastly, phase-level latency traces lay the groundwork for accurate runtime modeling, an essential capability for dynamic scheduling and improving interactive responsiveness in on-device LLM systems.

Unfortunately, directly profiling phase-level latency for on-device LLMs via application-level APIs is infeasible. First, application-level timing is inherently imprecise. User-facing apps typically interact with inference runtimes through high-level wrappers (e.g., Java/Kotlin on Android or Swift on iOS), which expose only coarse-grained timing signals, such as those from \texttt{SystemClock}, spanning the entire model dispatch. 
These measurements also include overheads from library calls, runtime transitions, and language bindings (e.g., JNI bridges to native C++), obscuring the true execution time. 
Second, application-level instrumentation lacks visibility into internal phase boundaries of LLM inference, which are handled within the native runtime.

{\sloppy
To enable accurate phase-level latency measurements, \textsc{lm-Meter} instruments the native inference engine, such as MLC~\cite{mlc}, at the runtime level. Specifically, we insert lightweight trace timers into the C++ backend, placed immediately before and after each semantic phase. Each timer is implemented using \texttt{std::chrono::steady\_clock}, a monotonic clock provided by the C++ \texttt{chrono} library, wrapped in a custom time utility to ensure consistent collection. 
We select \texttt{steady\_clock} after systematic evaluation, based on the following practical advantages:
(i) It is monotonic and immune to wall-clock adjustments, ensuring timestamp differences are reliable. (ii) It is supported across all major platforms (Linux, Android, iOS, macOS, and Windows), as required by the C++ standard, enabling portable instrumentation without platform-specific code.
(iii) On modern systems, \texttt{steady\_clock} maps to high-resolution hardware clocks, typically offering sub-microsecond precision, sufficient for capturing LLM phase durations ranging from microseconds to hundreds of milliseconds.
(iv) The call to \texttt{steady\_clock::now()} incurs minimal overhead. Consequently, \texttt{steady\_clock} strikes an effective balance between portability, accuracy, and low instrumentation overhead, making it a robust choice for fine-grained latency profiling in resource-constrained environments.
Additionally, \textsc{lm-Meter} tags each timing record with its corresponding phase name and sequence index, enabling alignment with the input prompt sequence and per-token generation steps. This facilitates per-token phase-level latency analysis, providing fine-grained insights into the temporal behavior of LLM inference across the generation sequence.
\par
}


\subsection{Kernel-level Latency Profiling} 

A kernel typically serves as the fundamental unit of scheduling and execution in modern machine learning (ML) frameworks, especially on mobile and edge platforms~\cite{zhang2021nn}.
The execution latency of individual kernels is highly sensitive to hardware characteristics, including compute capacity, memory bandwidth, and cache hierarchy. As a result, kernel performance can vary significantly across devices and fluctuate even on the same device due to dynamic resource contention from background applications. Hence, accurate, online profiling of kernel-level latency is essential for identifying performance bottlenecks and enabling hardware-aware optimizations, which includes device-specific kernel tuning, dynamic scheduling strategies, all of which are crucial for maximizing efficiency in on-device LLM systems.

However, accurate kernel-level runtime latency profiling for on-device LLMs faces two challenges, recognized in the broader systems research community. 
First, inserting trace timers directly into individual GPU kernel source code is infeasible because most mobile GPU drivers and runtime libraries are proprietary and closed-source. 
Second, modern ML compilers, such as TVM \cite{tvm} and LiteRT \cite{liteRT}, translate a graph of high-level operators (e.g., \texttt{Dequantize} $\rightarrow$ \texttt{MatMul} $\rightarrow$ \texttt{Add} $\rightarrow$ \texttt{RMSNorm}) into fused GPU kernels. During this process they apply kernel-fusion passes that merge multiple adjacent operators into a single launch to minimize memory traffic and kernel-launch overhead. These fusion passes are driven by heuristic cost models, autotuning, or backend-specific templates but the resulting fusion strategies and kernel boundaries are opaque to users.

\begin{figure}[t]
\centering
\begin{lstlisting}[style=cppstyle,
  basicstyle=\scriptsize\ttfamily,
  keywordstyle=\color{teal},
  morekeywords={CL_PROFILING_COMMAND_QUEUED,
                  CL_PROFILING_COMMAND_SUBMIT,
                  CL_PROFILING_COMMAND_START,
                  CL_PROFILING_COMMAND_END}]
OPENCL_CALL(clGetEventProfilingInfo(
    ev, CL_PROFILING_COMMAND_QUEUED, sizeof(r->t_queued_ns),
    &r->t_queued_ns, nullptr));
OPENCL_CALL(clGetEventProfilingInfo(
    ev, CL_PROFILING_COMMAND_SUBMIT, sizeof(r->t_submit_ns),
    &r->t_submit_ns, nullptr));
OPENCL_CALL(clGetEventProfilingInfo(
    ev, CL_PROFILING_COMMAND_START, sizeof(r->t_start_ns),
    &r->t_start_ns, nullptr));
OPENCL_CALL(clGetEventProfilingInfo(
    ev, CL_PROFILING_COMMAND_END, sizeof(r->t_end_ns),
    &r->t_end_ns, nullptr));
\end{lstlisting}
\caption{Code snippet for querying OpenCL kernel‐execution timestamps.}
\label{lst:opencl-timing}
\end{figure}

{\sloppy
To address these challenges, \textsc{lm-Meter} leverages the event callback mechanisms defined in GPU libraries, such as OpenCL, VulKan, and Metal, to capture the complete lifecycle of every kernel during model execution, without requiring access to kernel source code. Inspired by nnPerf \cite{nnPerf}, this approach enables fine-grained kernel-level profiling even in closed-source environments. 
Using OpenCL as an example, the runtime maintains three queues, the \textit{command queue}, \textit{host queue}, and \textit{device queue}. 
After model compilation, the interpreter constructs a command queue, where each entry represents either a GPU kernel invocation or a data movement operation. During inference, each entry from the command queue are copied to the host queue (CPU) and finally enqueued into the device queue (GPU) chronologically for execution. 
\textsc{lm-Meter} instruments this process by attaching profiling callbacks to each OpenCL command via the native \texttt{cl\_event} interface. Specifically, it queries the \texttt{cl\_event} object associated with each entry using \texttt{clGetEventProfilingInfo}, which exposes four standardized profiling nanosecond-resolution timestamps: 
\begin{itemize}
    \item \texttt{CL\_PROFILING\_COMMAND\_QUEUED}: when the command (e.g., a kernel launch or data movement) is enqueued into the command queue;
    \item \texttt{CL\_PROFILING\_COMMAND\_SUBMIT}: when the command is submitted by the host to the device queue; 
    \item \texttt{CL\_PROFILING\_COMMAND\_START}: when the device actually begin execution of the command;
    \item \texttt{CL\_PROFILING\_COMMAND\_END}: when the device complete execution of the command.
\end{itemize}
A code snippet illustrating this is shown in Fig. \ref{lst:opencl-timing}.
Additionally, \textsc{lm-Meter} maintains a lightweight per-kernel data structure containing: the kernel name, a pointer to its associated command queue, a host-side enqueue timestamp  \texttt{t\_cpu\_enqueue\_ns}, and the four device-side timestamps (\texttt{t\_queued\_ns}, \texttt{t\_submit\_ns}, \texttt{t\_start\_ns}, and \texttt{t\_end\_ns}). 
These measurements allow us to reconstruct the full execution timeline for each kernel launch during inference. They also enable precise attribution of latency across three key stages: (i) host-to-device queuing, (ii) scheduling and dispatch delays within the device runtime, and (iii) device-side kernel execution.
\par
}
\subsection{From Latency to Energy and Adaptation}
\textsc{lm-Meter} is designed with extensibility in mind and provides the infrastructure hooks to enable future work.

\textbf{Energy profiling.}
While \textsc{lm-Meter} currently targets latency analysis, its modular architecture readily supports integration with energy and power measurement tools, such as built-in current sensor on mobile devices or external power monitor like Monsoon \cite{Monsoon}. Time-aligned phase- and kernel-level traces make it a solid foundation for profiling on-device LLM energy and for joint latency–energy optimization.


\textbf{Runtime system adaptation.}
Although adaptation mechanisms are not implemented in this study, \textsc{lm-Meter} can serve as a telemetry backend for adaptive LLM systems. Real-time profiling signals can inform an \textit{Adaptation Engine} to dynamically adjust model selection, resource allocation, or execution strategies (e.g., model switching, frequency scaling, or phase skipping) in response to runtime conditions. By delivering structured latency feedback during inference, \textsc{lm-Meter} opens new opportunities for fine-grained, performance-aware adaptation in on-device LLMs.


\section{Evaluation}
\label{sc:evaluation}

In this section, we present a comprehensive evaluation of \textsc{lm-Meter}'s profiling accuracy and system overhead on commercial mobile devices. We begin by detailing the implementation of \textsc{lm-Meter}, followed by an in-depth analysis of its phase- and kernel-level profiling performance. We conclude with an assessment of the system overhead incurred. This evaluation is essential to demonstrate that \textsc{lm-Meter} provides reliable, high-fidelity measurements while remaining lightweight enough for real-time on-device LLM profiling.

\subsection{\textsc{lm-Meter} Implementation}
\label{ssc:implementation}
We implement \textsc{lm-Meter} on top of the MLC framework \cite{mlc} and TVM \cite{tvm}, comprising approximately 3,500 lines of C++ (e.g., instrumentation and profiler runtime), Python (e.g., model compilation, build tooling, and data post-processing), and Java/Kotlin (e.g., Android app) code to support the cross-stack nature of on-device LLM deployment, which span both back-end systems and front-end interfaces. Since OpenCL-based GPU acceleration is widely supported by mainstream ML frameworks, such as LiteRT (formerly known as TensorFlow Lite) \cite{liteRT}, ONNX \cite{onNX}, OpenVINO \cite{openvino} and others, \textsc{lm-Meter} is designed to be readily portable to alternative frameworks. While implementation details may differ across frameworks, the core profiling and instrumentation logic can be adapted with minimal effort. 

To ensure experimental repeatability, we fix the input prompts and set the maximum number of generation tokens.  
We configure the sampling parameters to \texttt{temperature = 0.0 and top\_p = 1.0}, and disable stop strings to enforce deterministic token generation during profiling. Unless otherwise specified, all experiments are repeated three times, and we report the average results.
All experimental results presented in Sections~\ref{sc:evaluation} and~\ref{sc:empirical} are obtained using three commercial mobile platforms, summarized in Table~\ref{tb:devices}. 
We present selected device combinations in each section to balance hardware diversity within space constraints.

\begin{table}[t]
\centering
\caption{Mobile devices used in our experiments.}
\label{tb:devices}
\begin{adjustbox}{max width=\columnwidth}
\begin{threeparttable}
\begin{tabular}{l|l|c|c|c}
\toprule[1pt]\midrule[0.3pt]
Devices & SoCs & RAM & Year  & Tier\\ \hline
Google Pixel 8 Pro & Google Tensor G3 & 12GB  & 2024  & High \\ 
Google Pixel 7 & Google Tensor G2 & 8GB & 2023  & Mid\\ 
Google Pixel 6 & Google Tensor & 8GB & 2022  & Low\\ 
\midrule[0.3pt]\bottomrule[1pt]
\end{tabular}
\end{threeparttable}
\end{adjustbox}
\end{table}

\textbf{Proposed profiling evaluation metrics.} To rigorously assess the profiling accuracy, we define two complementary metrics for each phase or kernel $k$:

\begin{enumerate}
\item \textit{Accuracy \(\alpha_k\)} (\%):
\begin{small}
\begin{equation}
\label{eq:abs}
      \alpha_k
      \;=\;
      \Bigl(1 - \frac{\bigl\lvert\,t_k^{\mathrm{LM}} - t_k^{\mathrm{GT}}\,\bigr\rvert}
                     {\,t_k^{\mathrm{GT}}}\Bigr)
      \times 100,
\end{equation}
\end{small}\noindent
where \(t_k^{\mathrm{LM}}\) and \(t_k^{\mathrm{GT}}\) denote the latency measured by \textsc{lm-Meter} and the ground-truth latency, respectively, in milliseconds. \(\alpha_k\) quantifies the proportion of the true execution time correctly captured by \textsc{lm-Meter}.

\item \textit{Scaled error rate \(\varepsilon_k^\star\)} (\(\mu\mathrm{s}/\mathrm{ms}\)):
\begin{small}
\begin{equation}
\label{eq:error‐rate}
\varepsilon_k^\star
    \;=\;
    \frac{10^3\,\Delta_k\,[\mu\mathrm{s}]}{\,t_k^{\mathrm{GT}}\,[\mathrm{ms}]}
    \quad(\mu\mathrm{s}\,\text{per ms}),
\end{equation}
\end{small}\noindent
where $\Delta_k = \bigl\lvert\,t_k^{\mathrm{LM}} - t_k^{\mathrm{GT}}\,\bigr\rvert$. By normalizing the absolute error by the ground-truth runtime (in milliseconds), \(\varepsilon_k^\star\) measures the number of microseconds of error incurred per millisecond of actual execution time. This unitless quantity is typically on the order of \(\mathcal{O}(1)\), offering an intuitive sense of relative error magnitude.
\end{enumerate}

In summary, \(\alpha_k\) provides a direct measure of ground-truth coverage as a percentage (higher is better), while \(\varepsilon_k^\star\) quantifies the absolute timing error normalized by runtime (lower is better). Together, these metrics offer complementary relative and absolute perspectives on the fidelity of the profiler.

\subsection{Phase-Level Profiling Performance}
\label{ssc:phaseperf}
\textbf{Phase-level latency ground-truth.} To evaluate the accuracy of phase-level profiling, we first establish a trusted baseline reference using AGI, which does not natively support phase-aware latency measurement for LLM inference.
To enable AGI-based tracing for on-device LLM inference, we extend the MLC runtime with instrumentation support via the Perfetto Tracing SDK. Specifically, we define and insert phase-level annotations using the \texttt{TRACE\_EVENT()} macro \cite{perfetto} to mark key inference phases, such as embedding, prefill, decoding, softmax, sampling, and probability transfer to CPU. All events are grouped under a unified tracing category (e.g., ``llm''), which is statically registered during runtime initialization.
During profiling, the target mobile device is connected via USB to a host machine running AGI. Upon completion of inference tasks, AGI generates a binary \texttt{.perfetto} trace file, which we post-process to extract timestamped phase events. In parallel, \textsc{lm-Meter} captures phase-level timing data in real time, enabling a direct comparison against the AGI measurements.


\begin{figure*}[t]
\centering
\subfigure[Single \emph{prefill} step. (Total absolute error $=0.2739\mathrm{ms}$)]
{\includegraphics[width=0.49\textwidth]{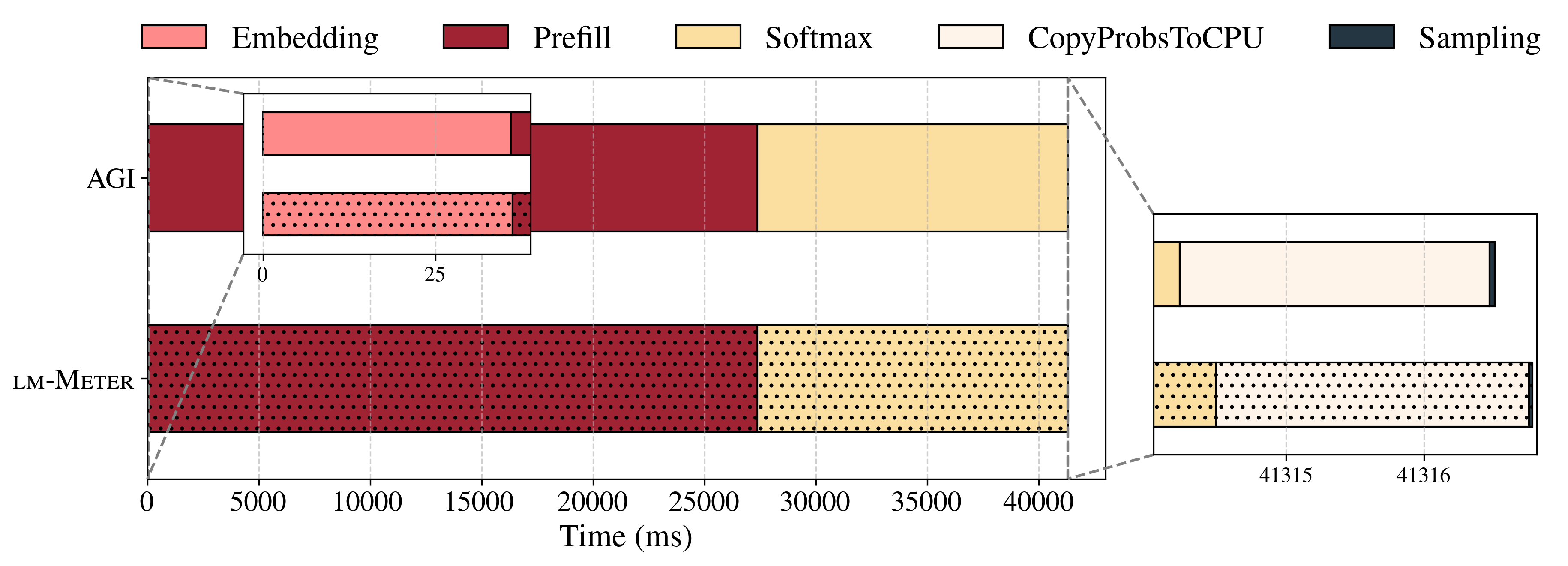}\label{fig:prefill}}
\subfigure[Three consecutive \emph{decode} steps. (Total absolute error $=0.1643\mathrm{ms}$)]
{\includegraphics[width=0.49\textwidth]{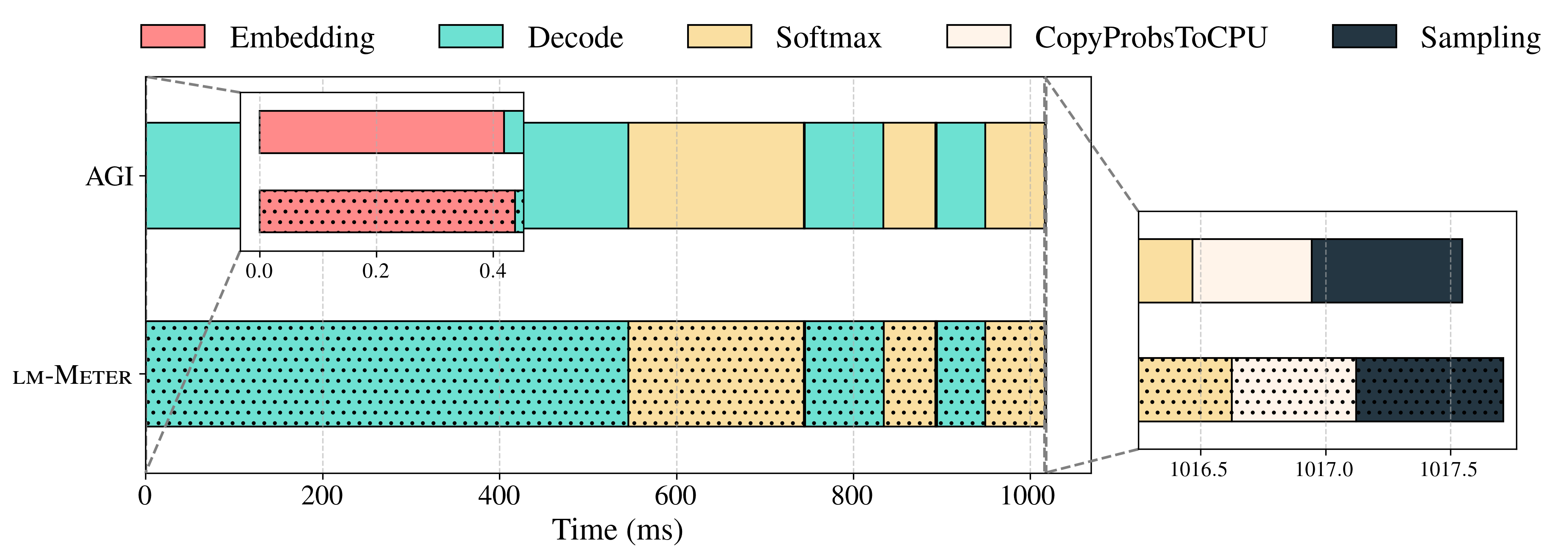}\label{fig:decode}}
\caption{Phase‐level latency comparison between our \textsc{lm-Meter} and AGI ground-truth on Google Pixel 8 Pro running a quantized \texttt{gemma-2-2b-it} model (4 bit weights, 16 bit activations). Insets zoom in on the sub-millisecond segments, including \emph{Embedding}, \emph{CopyProbsToCPU}, and \emph{Sampling}.}
\label{fig:phaselvl-gemma2}  
\end{figure*}

\begin{table}[t]
\centering
\caption{Phase-level profiling accuracy on Pixel 8 Pro.}
\label{tb:phase-8pro}
\begin{adjustbox}{max width=\columnwidth}
\begin{threeparttable}
\begin{tabular}{l|c|cc|c|c}
\toprule[1pt]\midrule[0.3pt]
\multirow{2}{*}{Models} & \multirow{2}{*}{Phases} & \multicolumn{2}{c|}{Profiled latency ($\mathrm{ms}$)} &  \multirow{2}{*}{$\alpha$ (\%)} & \multirow{2}{*}{$\varepsilon^\star$(\(\mu\mathrm{s}/\mathrm{ms}\))} \\\cline{3-4}
&  & \textsc{lm-Meter} &  \textsc{AGI} &        &       \\
\hline
\multirow{7}{*}{\rotatebox[origin=c]{90}{%
    \shortstack{\texttt{Llama-3.2-}\\\texttt{3B-Instruct}}}} & Embedding      & 0.8038 & 0.7763 & 96.46 & 35.412 \\
                                        & Prefill        & 3433.8628 & 3433.8142 & 99.99 & 0.014  \\
                                        & Decode         & 62.5669 & 62.5303 & 99.94 & 0.585  \\
                                        & Softmax        & 142.6166 & 142.6542 & 99.97  & 0.264  \\
                                        & CopyProbsToCPU & 0.4929 & 0.4616 & 93.22  & 67.718  \\
                                        & Sampling       & 0.0675 & 0.0824 & 81.86  & 181.439  \\\cline{2-6}
                                        & \textbf{End-to-end} &3640.4104&3640.3191&\textbf{99.99}&\textbf{0.025}\\
\hline
\multirow{7}{*}{\rotatebox[origin=c]{90}{%
    \shortstack{\texttt{Gemma-2-}\\\texttt{2B-it}}}} & Embedding      & 0.7659 & 0.7398 & 96.48 & 35.226 \\
                                        & Prefill        & 9301.1318 & 9301.0589 & 99.99 & 0.008  \\
                                        & Decode         & 54.5909 & 54.5557 & 99.94 & 0.646  \\
                                        & Softmax        & 502.3319 & 502.3698 & 99.99  & 0.076  \\
                                        & CopyProbsToCPU & 0.5570 & 0.5255 & 94.02  & 59.829  \\
                                        & Sampling       & 0.1698 & 0.1830 & 92.76  & 72.365  \\\cline{2-6}
                                        & \textbf{End-to-end} &9859.5473&9859.4329&\textbf{99.99}&\textbf{0.012}\\
\hline 
\multirow{7}{*}{\rotatebox[origin=c]{90}{%
    \shortstack{\texttt{TinyLlama-}\\\texttt{1.1B-Chat-}\\\texttt{v1.0}}}} & Embedding   &0.6858&0.6636&96.66&33.428\\
                                        & Prefill &5023.1182&5023.0765&99.99&0.008\\
                                        & Decode &34.6845&34.6534&99.91&0.897\\
                                        & Softmax &188.1463&188.1779&99.98&0.168\\
                                        & CopyProbsToCPU &0.3887&0.3602&92.10&78.980\\
                                        & Sampling &0.0504&0.0669&75.36&246.372\\\cline{2-6}
                                        & \textbf{End-to-end} &5247.0738&5246.9985&\textbf{99.99}&\textbf{0.014}\\
\hline 
\multirow{7}{*}{\rotatebox[origin=c]{90}{%
    \shortstack{\texttt{DeepSeek-}\\\texttt{R1-Distill-}\\\texttt{Qwen-1.5B}}}} & Embedding   &0.6753&0.6531&96.60&33.975\\
                                        & Prefill                     &7630.7553&7630.6979&99.99&0.008   \\
                                        & Decode                      &49.3840&49.3499&99.93&0.690\\
                                        & Softmax                     &437.0047&437.0459&99.99&0.094\\
                                        & CopyProbsToCPU              &0.4206&0.3888&91.84&81.565\\
                                        & Sampling                    &0.0671&0.0786&85.41&145.949   \\\cline{2-6}
                                        & \textbf{End-to-end}     &8118.3069&8118.2143&\textbf{99.99}&\textbf{0.011}\\
\hline 
\multirow{7}{*}{\rotatebox[origin=c]{90}{%
    \shortstack{\texttt{SmolLM2-}\\\texttt{360M-Instruct}}}} & Embedding   &0.9844&0.9671&98.21&17.909\\
                                        & Prefill &2399.3860&2399.3246&99.99&0.026\\
                                        & Decode &44.7612&44.7377&99.95&0.527\\
                                        & Softmax &221.8416&221.8672&99.99&0.116\\
                                        & CopyProbsToCPU &0.3081&0.2854&92.05&79.532\\
                                        & Sampling &0.0300&0.0394&76.06&239.395\\\cline{2-6}
                                        & \textbf{End-to-end} &2667.3113&2667.2214&\textbf{99.99}&\textbf{0.034}\\
\midrule[0.3pt]\bottomrule[1pt]
\end{tabular}
\begin{tablenotes}[flushleft]\small
\item[*] Latencies are reported to $10^{-4}$\,ms (0.1\,$\mu$s); 
  error‐rates $\varepsilon^\star$ are rounded to $10^{-3}$\,$\mu$s/ms; 
  accuracies $\alpha$ to $0.01$\,\%.
\item[$\dagger$] \texttt{SmolLM2-360M-Instruct} remains in full FP16 precision (unquantized), while all other models employ group-wise 4-bit weight quantization with 16-bit activations.
\end{tablenotes}
\end{threeparttable}
\end{adjustbox}
\end{table}

\textbf{Evaluation of phase-level profiling accuracy.}
Table~\ref{tb:phase-8pro} presents a comparison of phase-level runtime latencies measured by \textsc{lm-Meter} against baseline values obtained using AGI across five LLMs. These models span a wide range of scales, from 360 million to 3 billion parameters, and represent a diverse set of architectural designs. Among them, \texttt{SmolLM2-360M-Instruct} operates entirely in FP16 precision (unquantized), while the remaining models employ group-wise 4-bit weight quantization with 16-bit activations.
Each experiment is repeated three times, with ten prompt-response turns per run. We report the mean latency in Table \ref{tb:phase-8pro}. 
Across all models and phases, \textsc{lm-Meter} consistently demonstrates high profiling accuracy, achieving end-to-end latency accuracy\footnote{End-to-end latency is computed as the sum of all phase latencies.} of $\alpha \geq 99.99 \%$ and scaled error \(\varepsilon_k^\star\) $\leq$ 0.034 \(\mu\mathrm{s}/\mathrm{ms}\), indicating that \textsc{lm-Meter} introduces only negligible deviation from AGI measurements. 

Additionally, inference phases with the greatest impact on end-to-end latency, including prefill, decode, and softmax, are profiled with exceptional accuracy. These phases account for over $99\%$ of total inference time, as shown in Fig. \ref{fig:phaselvl-gemma2}, and exhibit profiling accuracy of $\alpha \geq 99.91 \%$ and \(\varepsilon_k^\star\) $\leq$ 0.897 \(\mu\mathrm{s}/\mathrm{ms}\).
The only notable outlier is the sampling phase. On the Pixel 8 Pro, a single sampling phase typically completes within just $50$--$80 \mu\mathrm{s}$, making it highly sensitive to profiling noise. Consequently, the inherent $\pm1$--$3$ $\mu\mathrm{s}$ timing variability of \textsc{lm-Meter} constitutes a substantial fraction of the total phase duration. This results in noticeably reduced profiling accuracy ($\alpha = 75$--$92\%$) and elevated scaled error (\(\varepsilon_k^\star\) $\leq$ 246 \(\mu\mathrm{s}/\mathrm{ms}\)).
Note that this reduced accuracy is not caused by limitations in the profiler’s granularity. Rather, it likely result from the inherent difficulty of measuring an \textit{extremely short}, partially host-bound routine whose execution time is comparable to system-level noise, such as OS scheduling jitter and CPU-GPU synchronization delays.

In summary, these results collectively demonstrate that \textsc{lm-Meter} delivers accurate and reliable phase-level profiling across a wide range of model scales and architectures.

\begin{table*}[t]
\centering
\caption{Kernel-level runtime latency profiling results on Google Pixel 8 Pro and Pixel 7.}
\begin{adjustbox}{max width=2\columnwidth,center}
\begin{threeparttable}
\begin{tabular}{l|c|c|c|c|c||c|c}
\toprule[1pt]\midrule[0.3pt]
\multirow{3}{*}{Kernels} & \multirow{3}{*}{Phases} & \multicolumn{4}{c||}{Google Pixel 8 Pro}   &   \multicolumn{2}{c}{Google Pixel 7} \\\cline{3-8}

& & \multicolumn{2}{c|}{Profiled latency ($\mathrm{ms}$)} &  \multirow{2}{*}{$\alpha$ (\%)} & \multirow{2}{*}{$\varepsilon^\star$(\(\mu\mathrm{s}/\mathrm{ms}\))} &  \multirow{2}{*}{$\alpha$ (\%)} & \multirow{2}{*}{$\varepsilon^\star$(\(\mu\mathrm{s}/\mathrm{ms}\))} \\\cline{3-4}
&  & \textsc{lm-Meter} &  GT &        &    &    &   \\
\hline

\texttt{dequantize1\_NT\_matmul5} & \multirow{4}{*}{Prefill} & 81.1899   & 82.1329  &  98.85  & 11.481 &98.88&11.212\\
\texttt{dequantize2\_NT\_matmul6} &                          & 31.3407   & 31.7568  &  98.69  & 13.103 &95.18&48.209\\
\texttt{dequantize3\_NT\_matmul7} &                          & 330.3757  & 332.7218  & 99.29  & 7.051 &98.87&11.328\\
\texttt{dequantize4\_NT\_matmul8} &                          & 367.5603  & 367.0284  & \textbf{99.86 (highest)}  & 1.449 &99.11&8.896\\
\hline
\texttt{dequantize1\_NT\_matmul10} & \multirow{9}{*}{Decode} & 0.3643  &  0.3737 &  97.46  & 25.391  &  97.19  &  28.145 \\
\texttt{dequantize2\_NT\_matmul11} & & 0.2062 & 0.2006  & 97.23  & 27.706 &98.14&18.587\\
\texttt{dequantize3\_NT\_matmul12} & & 1.3813 & 1.3601  & 98.44  & 15.587  &98.17&18.267\\
\texttt{dequantize4\_NT\_matmul13} & & 0.6862 & 0.6586  & 95.81  & 41.921   &97.50&25.044\\
\texttt{dequantize\_NT\_matmul14\_divide2\_tir\_tanh2\_multiply8} & & 18.4379 & 18.3619  & 99.59 & 4.147  &98.13&18.705\\
\texttt{add\_norm\_prefill} & & 0.1149 & 0.1059  & \textbf{91.51(lowest)}  & 84.891   &93.29&67.080\\
\texttt{rms\_norm2}         & & 0.1037 & 0.1092  & 94.93  & 50.641  &92.65&73.531\\
\texttt{split2\_gelu\_tanh2\_multiply7} & & 0.0952 & 0.0939 & 98.62 & 13.727  &93.75&62.517\\
\texttt{multiply6}        & & 0.1061 & 0.1005  & 94.35  & 56.546  &\textbf{90.31}&96.934\\
\hline
\texttt{chunk\_lse} & \multirow{2}{*}{Softmax}           & 0.2718 & 0.2839 & 95.53  & 44.735  &99.39&6.026\\
\texttt{softmax\_with\_chunked\_sum} &  & 0.2376 & 0.2392  &  99.33  & 6.689  &\textbf{99.40} &5.992\\
\hline
\texttt{dequantize\_take1} & Embedding & 0.1034 & 0.1097 & 94.26  & 57.429  &95.73&42.676\\
\midrule[0.3pt]\bottomrule[1pt]
\end{tabular}
\begin{tablenotes}[flushleft]\small
\item[*] Latencies are reported to $10^{-4}$\,ms (0.1\,$\mu$s); error‐rates $\varepsilon^\star$ are rounded to $10^{-3}$\,$\mu$s/ms; accuracies $\alpha$ to $0.01$\,\%.
\item[$\dagger$] Model: \texttt{Gemma-2-2B-it} with group-wise 4-bit weight quantization with 16-bit activations.
\end{tablenotes}
\end{threeparttable}
\end{adjustbox}
\label{tb:kernel-8pro}
\end{table*}

\subsection{Kernel-Level Profiling Performance}
\label{ssc:kernelperf}

\textbf{Kernel-level latency groundtruth.} Due to the lack of existing profilers capable of accurately capturing kernel-level latency for on-device LLMs (\S\ref{ssc:profilers}), we estimate ground-truth latencies using a custom \textit{kernel duplication} technique.
Specifically, for each target kernel within an LLM inference phase, we perform the following two-step profiling:
\begin{itemize}
    \item Baseline profiling: We first profile the \textit{unmodified} model using AGI, recording the latency of each inference phase as $t_{i}$, where $i$ indexes the phases (e.g., embedding, prefill, decode, softmax).
    \item Kernel-duplicated profiling: We compile a modified model where the target kernel is duplicated $n$ times and inserted immediately after its original invocation within the same phase. Profiling this modified model yields updated phase latencies $t^{dup}_{i}$.
\end{itemize}
We estimate the ground-truth latency of the target kernel as: $\frac{t^{dup}_{i} - t_{i}}{n}$. We set $n=50$ for kernels with expected execution latency exceeding $1 \mathrm{ms}$, and $n=1000$ for shorter kernels. This procedure is repeated independently for each kernel of interest to obtain reliable per-kernel latency estimates.

\begin{figure}[t]
\centerline
{\includegraphics[width=0.48\textwidth]{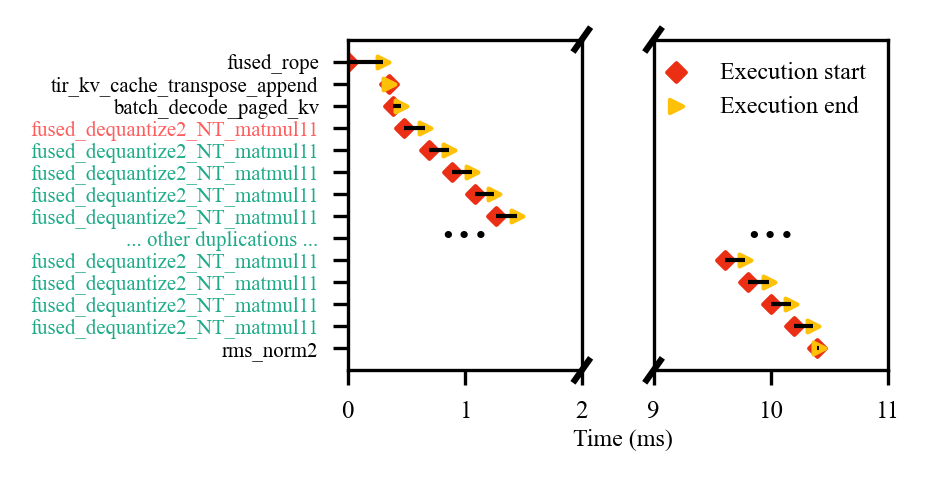}}
\caption{Illustration of kernel duplication for \texttt{fused\_dequantize2\_NT\_matmul11}, showing the original kernel (red) and its 50 duplicated instances (green). A broken x-axis with omitted rows is used to save space.}
\label{fig:kernel-dup}  
\end{figure}

{\sloppy
To implement the kernel duplication technique without introducing extra overhead or altering the semantics of the model, we modify the model compilation pipeline in MLC, which includes the code generation (codegen) path for kernel emission. Specifically, (1) the kernel generation is extended to emit $n$ back-to-back copies of target kernel, preserving the original data dependencies; (2) optimization passes that may eliminate unused or redundant code, such as \texttt{remove\_all\_unused} and \texttt{DeadCodeElimination} are explicitly disabled to retain the duplicates; and (3) CPU and GPU frequencies are fixed to eliminate variability from dynamic voltage and frequency scaling (DVFS), and all non-essential foreground/background applications are terminated to minimize system-level noise.
Fig. \ref{fig:kernel-dup} shows kernel duplication for \texttt{fused}\texttt{\_dequantize2}\texttt{\_NT\_}\texttt{matmul11} in decode phase. 
\par
}

\textbf{Evaluation of kernel-level profiling accuracy.} 
Table~\ref{tb:kernel-8pro} presents kernel-level latency measurements captured by \textsc{lm-Meter} on both the Google Pixel 8 Pro and Pixel 7, compared against ground-truth (GT) values collected via AGI with kernel duplication. Across both platforms, \textsc{lm-Meter} consistently delivers high profiling accuracy at kernel granularity.
Specifically, on the Pixel 8 Pro, the mean and median accuracy across all 16 evaluated kernels are 96.82\% and 97.35\%, respectively. On the Pixel 7, the corresponding values are 96.61\% and 97.81\%. Even in the worst-case scenario, \texttt{multiply6}, \textsc{lm-Meter} achieves 90.31\% accuracy.
The four GEMM kernels within the prefill phase, which dominate the total inference latency, are profiled with near-perfect accuracy. On the Pixel 8 Pro, these kernels achieve \SIrange{98.69}{99.86}{\%} accuracy with scaled error rates $\varepsilon^\star<\SI{13.1}{\micro\second\per\milli\second}$; they reach \SIrange{95.18}{99.11}{\%} accuracy with $\varepsilon^\star<\SI{48.2}{\micro\second\per\milli\second}$ on the Pixel 7.
Moreover, \textsc{lm-Meter} also demonstrates high accuracy on \textit{micro-kernels} with true runtimes under \SI{1}{\milli\second}. On the Pixel 8 Pro, these kernels achieve a mean accuracy of \SI{95.44}{\%} and a median of \SI{95.23}{\%}; on the Pixel 7, the mean and median accuracy are \SI{95.73}{\%} and \SI{96.46}{\%}, respectively. 

In summary, these results validate \textsc{lm-Meter}'s reliability in accurately capturing the latency of both latency-dominant kernels and short-duration micro-kernels. 

\begin{figure*}[t]
\centering
\subfigure[CPU governor set to \texttt{Performance}]
{\includegraphics[width=0.32\textwidth]{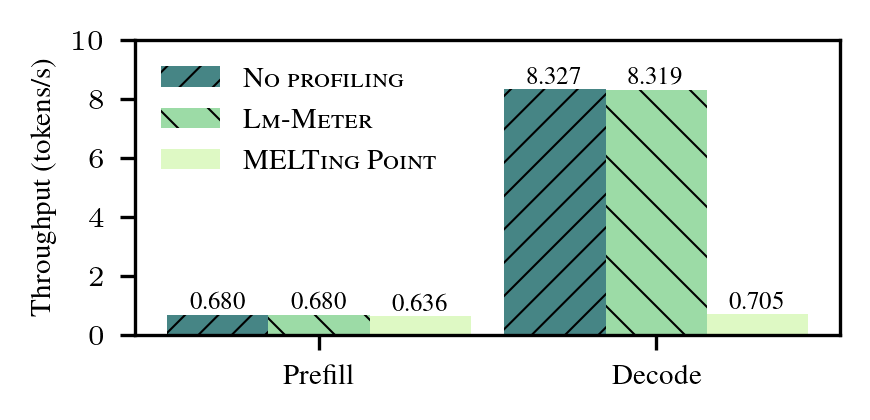}\label{fig:overhead_performance}}
\subfigure[ CPU governor set to \texttt{Conservative}]
{\includegraphics[width=0.32\textwidth]{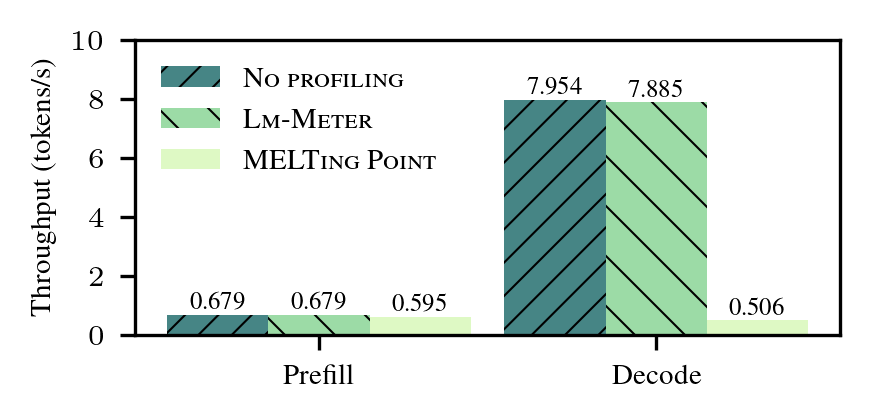}\label{fig:overhead_conservative}}
\subfigure[CPU governor set to \texttt{Powersave}]
{\includegraphics[width=0.32\textwidth]{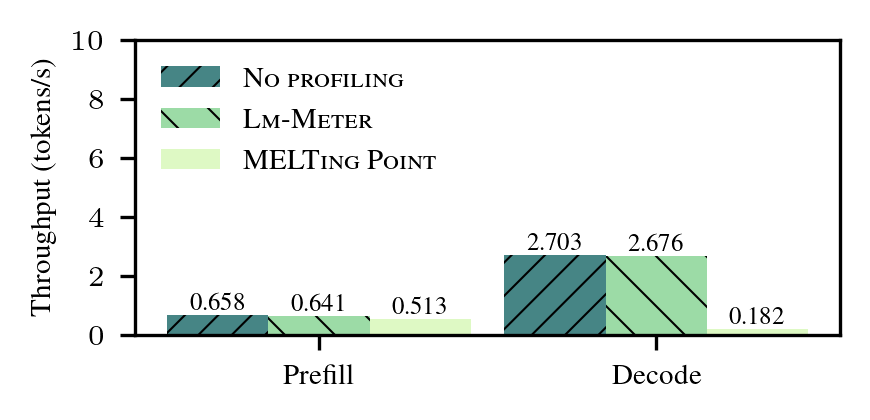}\label{fig:overhead_powersave}}
\caption{Comparison of profiling-induced latency overhead, measured as throughput (tokens/s) during prefill and decode, under three CPU governors. Results compare \textsc{lm-Meter}, MELTing Point~\cite{laskaridis2024melting}, and a no-profiling baseline.}
\label{fig:overhead_exam}  
\end{figure*}

\subsection{System Overhead}
\label{ssc:overhead}
Since \textsc{lm-Meter} performs online profiling directly on mobile devices, its practicality depends critically on introducing minimal latency overhead. To quantify this overhead, we measure the throughput (in tokens per second) of the two primary inference phases, prefill and decode, under three CPU governor configurations that emulate varying levels of on-device resource availability: (1) \texttt{Performance}: All CPU cores are prioritized to operate at their peak frequencies; (2) \texttt{Conservative}: Dynamic DVFS with a bias toward lower frequencies; (3) \texttt{Powersave}: All CPU cores are prioritized to run at their minimum frequencies.

We compare \textsc{lm-Meter} against two baselines: (1) \textit{No profiling} and (2) MELTing Point~\cite{laskaridis2024melting}, a state-of-the-art profiler for on-device LLMs. 
Fig. \ref{fig:overhead_exam} summarizes the results. Across all CPU governors, \textsc{lm-Meter} consistently imposes negligible throughput degradation. Under the \texttt{Performance} and \texttt{Conservative} governors, its throughput closely matches that of the no-profiling baseline for both phases. Even under the \texttt{Powersave} setting, where system resources are most constrained, \textsc{lm-Meter} exhibits only a modest throughput reduction of $2.58\%$ for prefill and $0.99\%$ throughput for decode. 
In contrast, MELTing Point incurs substantial profiling overhead in all three settings, especially under \textit{Powersave}, where its throughput drops by $22\%$ for prefill and by more than $93\%$ for decode. 
These results underscore the efficiency of \textsc{lm-Meter} and affirm its suitability for real-time, online profiling in resource-constrained mobile environments.

\section{On-Device LLM Empirical Study}
\label{sc:empirical}
Through \textsc{lm-Meter}, we conduct two empirical studies: \textit{phase-oriented} and \textit{kernel-oriented}, to illuminate realistic performance behavior of on-device LLMs. 


\subsection{Phase-oriented Empirical Study}
\label{ssc:phase}
In this section, we present a phase-oriented empirical study of on-device LLMs and an in-depth analysis leveraging our \textsc{lm-Meter}. The goal is to characterize the trade-offs between model accuracy and efficiency during the \textit{prefill} and \textit{decode} phases across diverse tasks, model suites, and parameter scales.
To our knowledge, this is among the first systematic explorations of phase-level performance-efficiency trade-offs in LLM inference for on-device environments, which can contribute insightful suggestions for optimizing on-device LLM deployment and lay a foundation for future research.



\begin{table}[t]
\centering
\caption{Model suites for phase-oriented study.}
\label{tb:modelsuites}
\begin{adjustbox}{max width=\columnwidth}
\begin{threeparttable}
\begin{tabular}{l|l|c|c}
\toprule[1pt]\midrule[0.3pt]
Model suites & Model scales (parameters) & Precision  & HuggingFace collections \\
\hline
\texttt{Pythia}~\cite{biderman2023pythia} & \{70M, 160M, 410M, 1B, 1.4B\} & float16   & EleutherAI/pythia\\
\texttt{SmolLM2}~\cite{smolm2} & \{135M, 360M, 1.7B\} & bfloat16 &  HuggingFaceTB/SmolLM2\\
\texttt{Qwen1.5}~\cite{qwen} & \{0.5B, 1.8B\} & bfloat16 & Qwen/Qwen1.5\\
\midrule[0.3pt]\bottomrule[1pt]
\end{tabular}
\end{threeparttable}
\end{adjustbox}
\end{table}

\textbf{Model suite selection.} We evaluate three representative model suites: \texttt{Pythia}~\cite{biderman2023pythia}, \texttt{SmolLM2}~\cite{smolm2}, and \texttt{Qwen1.5}~\cite{qwen}, each offering a range of model sizes suited for affordable on-device deployment. Details are summarized in Table \ref{tb:modelsuites}.

\begin{table}[t]
\centering
\caption{Evaluation tasks for phase-oriented study.}
\label{tb:tasks}
\begin{adjustbox}{max width=\columnwidth}
\begin{threeparttable}
\begin{tabular}{l|l|c|c}
\toprule[1pt]\midrule[0.3pt]
Categories & Tasks & Full datasets  &  Sampled subsets  \\ \hline
\multirow{2}{*}{Knowledge ability} & ARC-Easy \cite{allenaiarc} & 2.38k prompts & 238 prompts \\
                                   & ARC-Challenge \cite{allenaiarc} & 1.12k prompts & 112 prompts\\
\hline
\multirow{2}{*}{Inference ability} & HellaSwag \cite{zellers2019hellaswag} & 10k prompts & 100 prompts \\
                                   & GSM8K \cite{cobbe2021gsm8k} & 1.32k prompts & 131 prompts\\
\midrule[0.3pt]\bottomrule[1pt]
\end{tabular}
\end{threeparttable}
\end{adjustbox}
\end{table}

\textbf{Task selection.} To capture a broad spectrum of model capabilities, we categorize mainstream evaluation tasks into two primary skill domains: \textit{knowledge ability} and \textit{reasoning ability}. 
Knowledge ability measures factual understanding without additional reasoning, while reasoning ability reflects multi-step inference grounded in prior knowledge.
In this study, we select two representative datasets for each category, listed in Table~\ref{tb:tasks}. ARC-Easy and ARC-Challenge \cite{allenaiarc} assess factual recall and basic comprehension, while HellaSwag \cite{zellers2019hellaswag} and GSM8K \cite{cobbe2021gsm8k} focus on reasoning. Specifically, HellaSwag evaluates commonsense reasoning in everyday contexts, and GSM8K targets multi-step mathematical problem solving.
These tasks vary in complexity and input length, offering a diverse view of inference behavior across different demands.

\begin{figure*}[t]
\centering
\subfigure[ARC-Easy (KL $=0.0195$)]
{\includegraphics[width=0.24\textwidth]{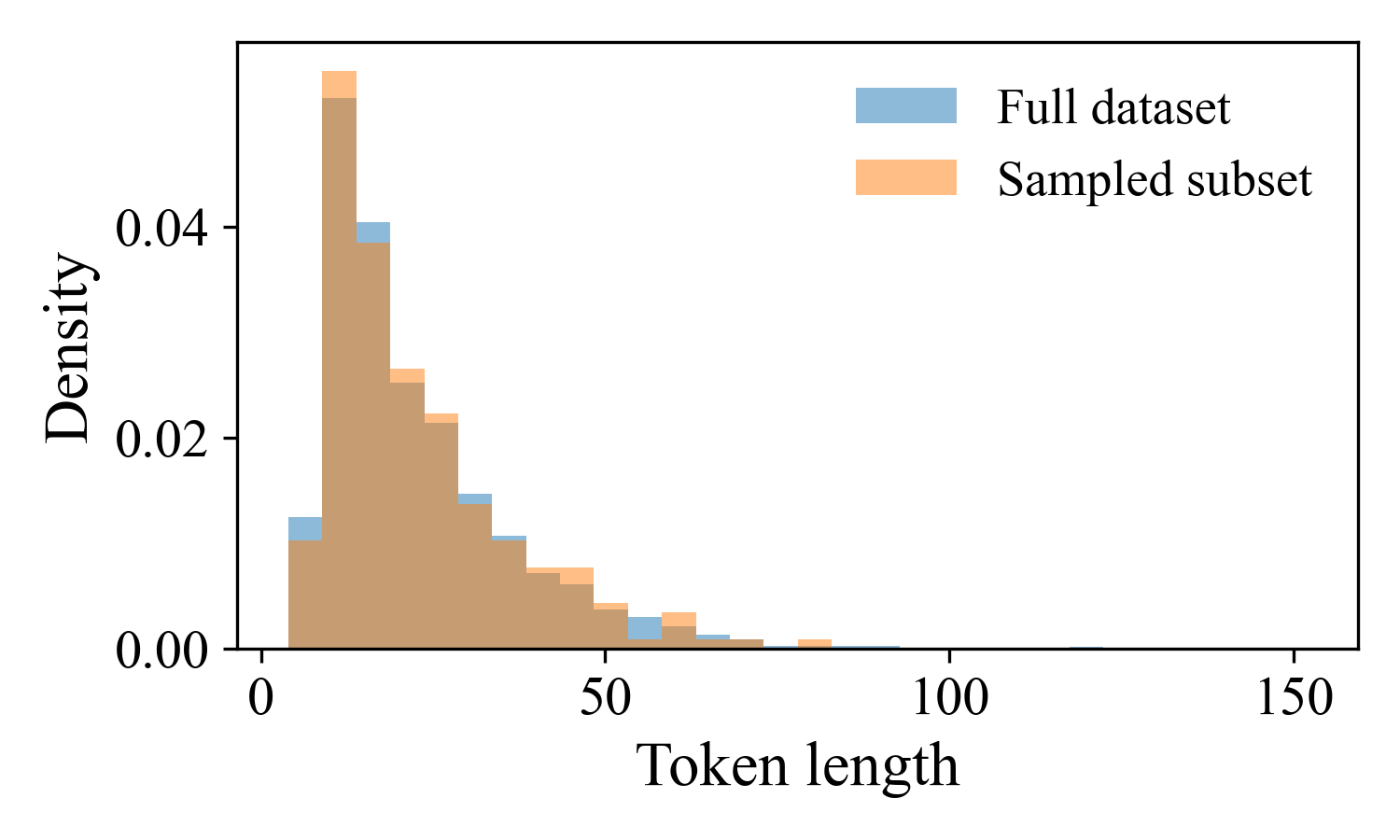}\label{fig:arc_easy}}
\subfigure[ARC-Challenge (KL $=0.0311$)]
{\includegraphics[width=0.24\textwidth]{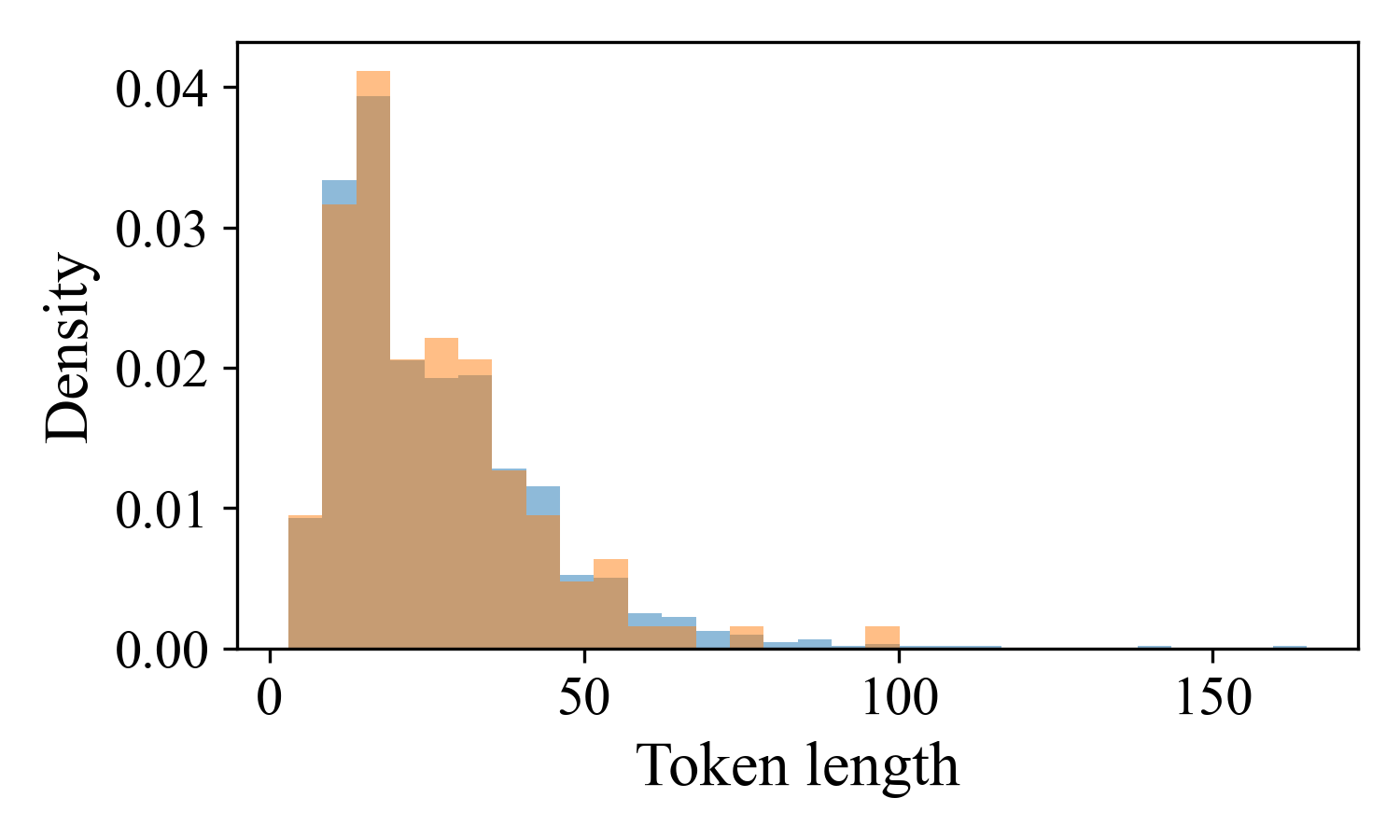}\label{fig:arc_challenge}}
\subfigure[HellaSwag (KL $=0.0433$)]
{\includegraphics[width=0.24\textwidth]{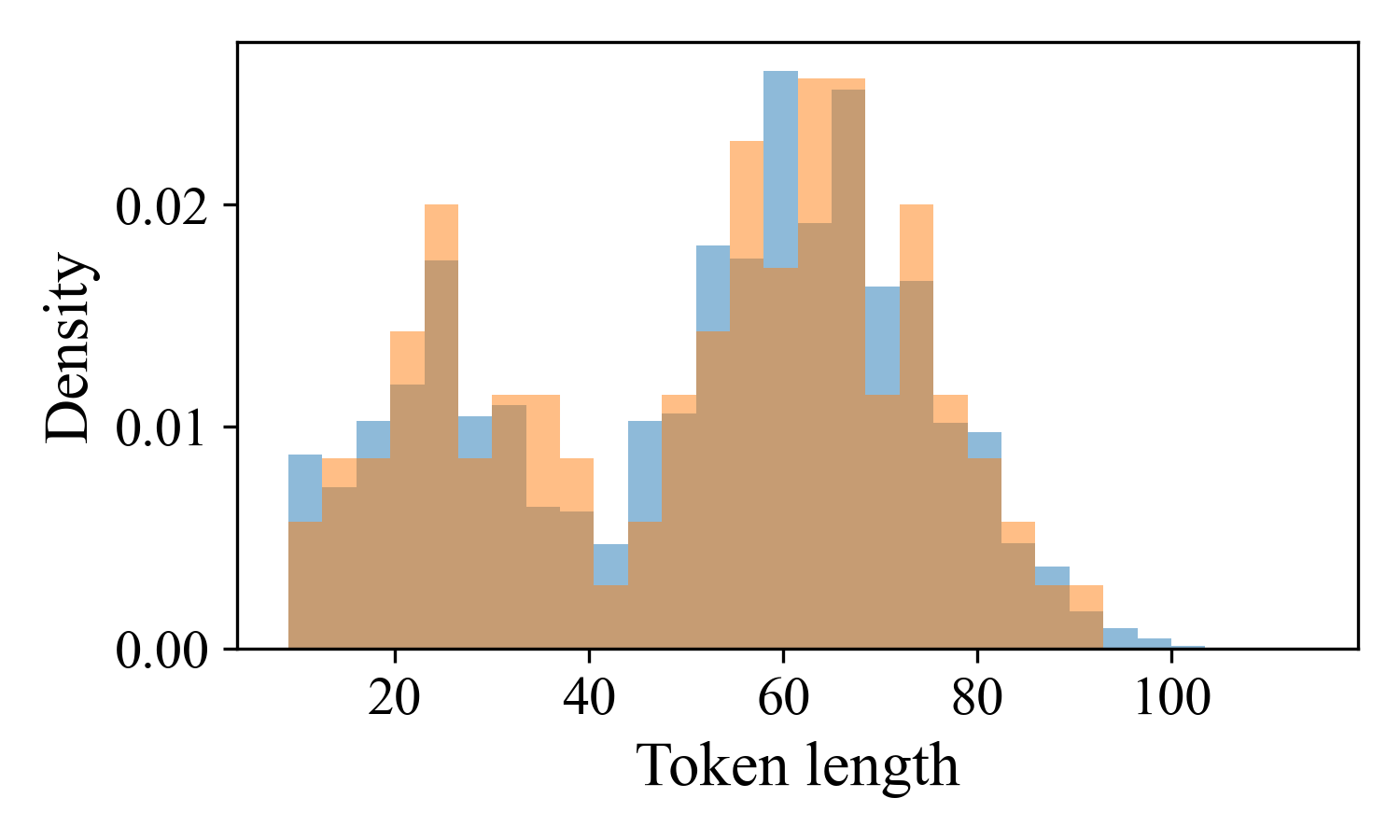}\label{fig:hellaswag}}
\subfigure[GSM8K (KL $=0.0330$)]
{\includegraphics[width=0.24\textwidth]{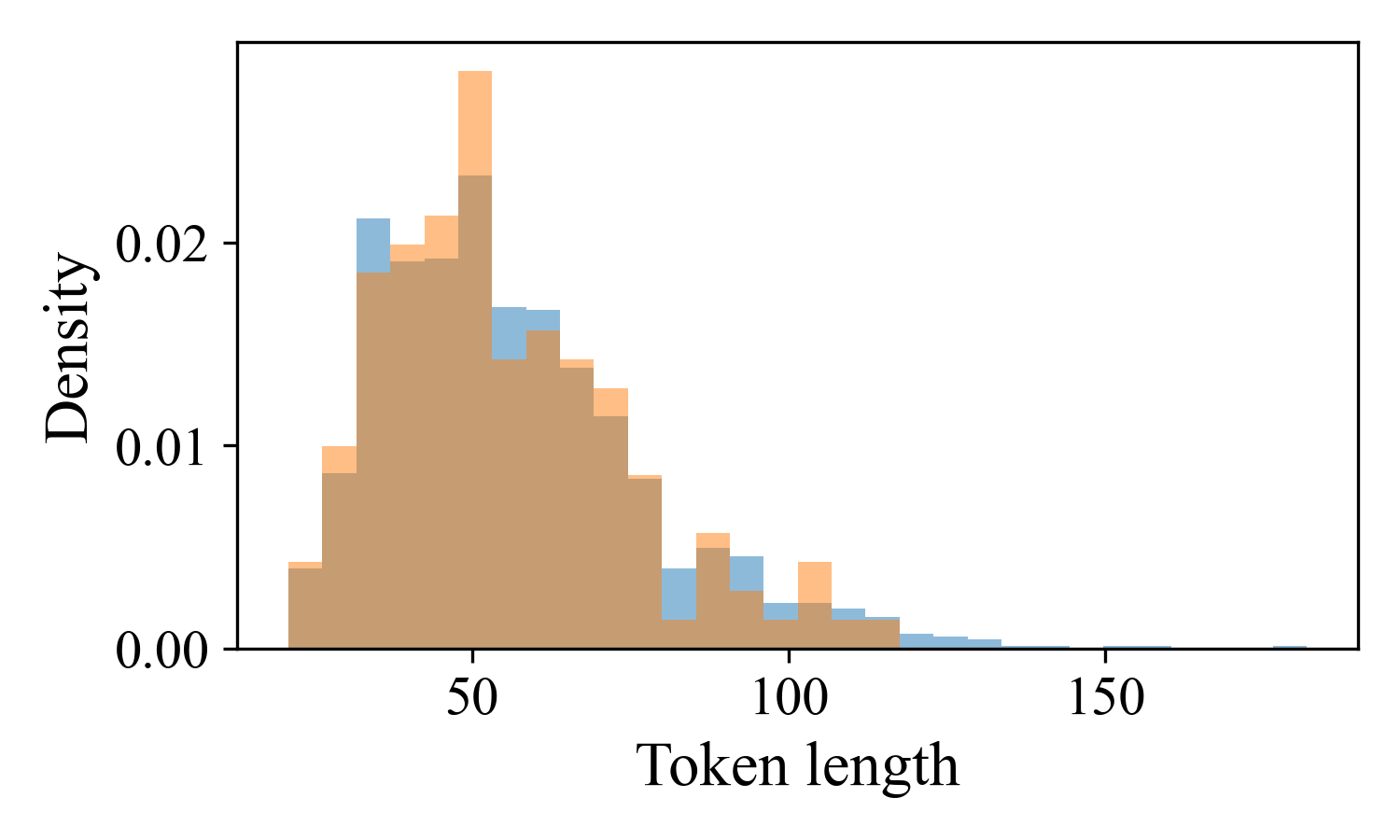}\label{fig:gsm8k}}
\caption{Token-length distributions of each full dataset (blue) vs. KL-matched subset (orange). We sample 10\% of ARC-Easy, ARC-Challenge, and GSM8K (238, 112, and 131 prompts) and 1\% of HellaSwag (100 prompts). Reported KL divergences measure how closely each subset’s token-length histogram matches the full distribution.}
\label{fig:subset}
\end{figure*}

\textbf{Subset sampling and evaluation.}
We find that full-benchmark evaluation of LLMs is often impractical under constrained resources. First, inference on mobile and edge hardware is significantly slower than on server-scale accelerators. For instance, running the full HellaSwag test set (10k prompts) can take hours or even days per model. As the number of models and tasks grows, evaluation cost scales poorly: $\text{Cost}=\# \text{models} \times \# \text{tasks} \times \# \text{prompts} \times \text{inference latency} \times \# \text{devices}$. Second, prolonged evaluation risks triggering thermal throttling or battery degradation on mobile devices, incurring inconsistent or unreliable measurements.

{\sloppy
To address the challenges of full-scale on-device evaluation, we develop a sampling-based approach. Instead of randomly selecting prompts, which may introduce bias due to skewed token-length distributions, we apply a Kullback-Leibler (KL) divergence-based sampling algorithm. This method selects a representative subset whose token-length distribution closely matches that of the full dataset, thereby preserving task complexity and linguistic diversity.
To balance scalability and fidelity, we retain 10\% of ARC-Easy, ARC-Challenge, and GSM8K (238, 112, and 131 prompts, respectively), and 1\% subset of HellaSwag (100 prompts), due to its larger size.
Fig.~\ref{fig:subset} illustrates the token-length distributions of both the full datasets and our sampled subsets. Notably, in all cases, the KL divergence remains below $0.0433$, demonstrating high representativeness.
Additionally, to streamline benchmarking across different models and tasks, \textsc{lm-Meter} automates prompt injection from the KL-matched subsets directly into the LLM inference pipeline. This removes manual interaction and eliminates human-induced variability, e.g., typing, ensuring consistent, repeatable, and unbiased runtime measurements.
\par
}

\begin{figure*}[t]
\centerline
{\includegraphics[width=0.98\textwidth]{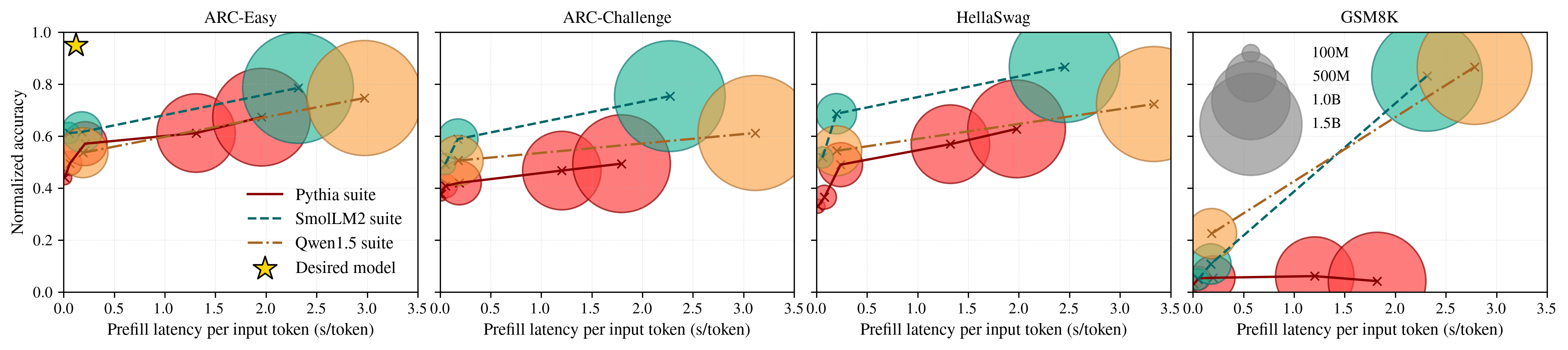}}
\caption{\textbf{Prefill} latency per input token vs. accuracy trade-off across LLM suites and tasks on \textbf{Pixel 8Pro}. Bubble area~$\propto$~model size; accuracy is normalized to LLaMA-33B~\cite{touvron2023llama} on each task for fair cross-benchmark comparison.} 
\label{fig:pref_speed_acc}
\end{figure*}
\begin{figure*}[t]
\centerline
{\includegraphics[width=0.98\textwidth]{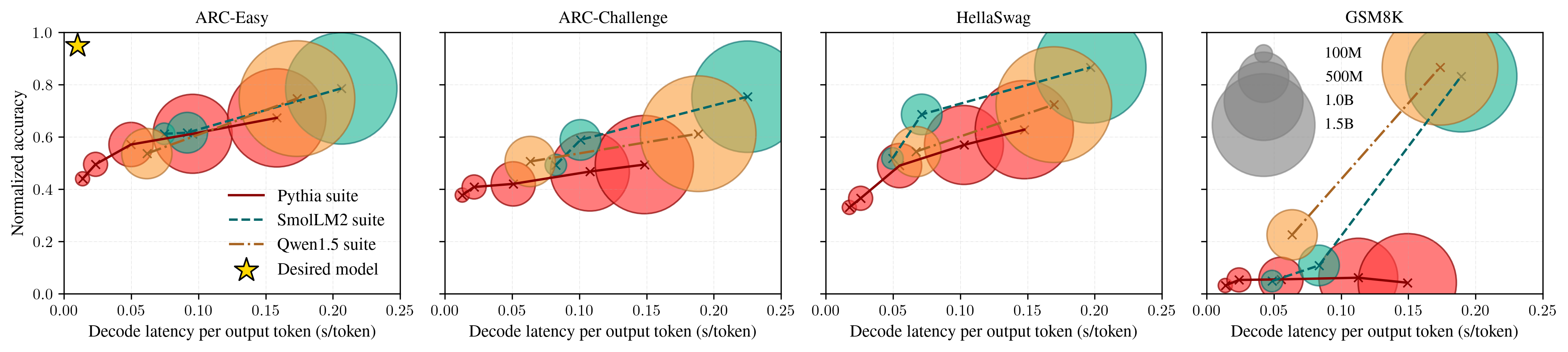}}
\caption{\textbf{Decode} latency per output token vs. accuracy trade-off across LLM suites and tasks on \textbf{Pixel 8Pro}.} 
\label{fig:dec_speed_acc}  
\end{figure*}
\begin{figure*}[t]
\centerline
{\includegraphics[width=0.98\textwidth]{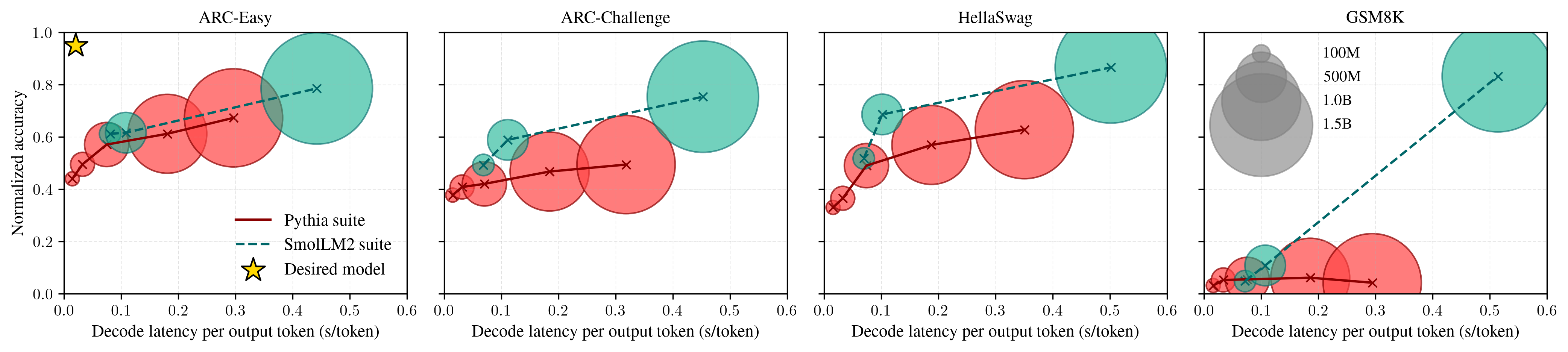}}
\caption{\textbf{Decode} latency per output token vs. accuracy trade-off across LLM suites and tasks on \textbf{Pixel 6}. \texttt{Qwen1.5-1.8B} is not evaluated due to insufficient memory on the device.} 
\label{fig:dec_speed_acc6}  
\end{figure*}

\textbf{Accuracy vs.\ decode/prefill latency.} We evaluate phase-specific efficiency using two metrics: \textit{decode latency per output token}\footnote{Decode latency per output token (s/token) measures how quickly the model generates tokens autoregressively during the decode phase.} and \textit{prefill latency per input token}\footnote{Prefill latency per input token (s/token) quantifies how efficiently the model processes the input context during the prefill phase.}. Model accuracy on individual tasks is evaluated using the \textit{lm-evaluation-harness}~\cite{evalharness}, while latency profiling is conducted with \textsc{lm-Meter}. To enable a fair and consistent comparison across tasks, the accuracy of each model is normalized to the performance of LLaMA-33B
~\cite{touvron2023llama} on the corresponding benchmark.
We perform the same experiments on both Pixel 8Pro and 6. The results are shown in Figs. \ref{fig:pref_speed_acc}, \ref{fig:dec_speed_acc}, and \ref{fig:dec_speed_acc6}, from which we derive the following key observations and implications.

First, \textit{prefill emerges as the dominant throughput bottleneck in on-device LLM inference, in contrast to trends observed in server-scale deployments.} 
For latency-sensitive, small-batch inference workloads common in interactive LLM applications, the autoregressive decode phase is widely recognized as the primary performance bottleneck on server-scale GPUs. Its inherently sequential execution and limited batching efficiency lead to significantly lower hardware utilization compared to the highly parallelizable prefill phase.
Prior work has demonstrated that prefill can be up to $100\times$ faster than decode on a single A100 GPU at batch size 1~\cite{agrawal2024taming}. 
However, our findings reveal the opposite trend on mobile and edge devices. For example, when scaling the \texttt{Pythia} from 70M to 1.4B parameters (at batch size 1), the average prefill latency per input token across four tasks increases dramatically from 0.012 s/token to 1.9 s/token, a $158\times$ slowdown. In contrast, decode latency per output token grows more moderately, from 0.015 s/token to 0.15 s/token, representing only a $10\times$ slowdown.
This inversion arises because the prefill requires processing the entire input context in a single forward pass, which becomes increasingly expensive on mobile and edge hardware with constrained compute resources, limited memory bandwidth, and the lack of support for parallelized prefill.

\begin{graybox}
\underline{Implication 1:} \textit{These observations underscore the need for phase-specific optimizations tailored to the unique constraints of on-device LLMs, particularly prefill. While decode-phase optimizations have been the primary focus in server-scale LLM inference, such as speculative decoding~\cite{leviathan2023fast,chen2023accelerating}, paged KV cache~\cite{kwon2023efficient}, and tensor parallelism~\cite{hansen2024communication}, their effectiveness in on-device settings remains under-explored.}
\end{graybox}

Second, \textit{accuracy-latency trade-offs in on-device LLMs are highly task-dependent.} We observe that the extent to which a model can balance inference quality (accuracy) and efficiency (latency) varies significantly across tasks, likely due to differences in input complexity, prompt structure, and reasoning requirements.
As illustrated in Fig.~\ref{fig:dec_speed_acc} and \ref{fig:dec_speed_acc6}, for instance, ARC-Easy comprises relatively simple multiple-choice questions with short, factual prompts. In this case, reducing model size significantly improves latency without incurring a steep accuracy penalty. Compared to \texttt{Pythia-1.4B}, the average normalized accuracies of \texttt{Pythia-1B}, \texttt{410M}, \texttt{160M}, and \texttt{70M} degrade by only 9\%, 15\%, 27\%, and 34\%, respectively, while their decode latency per output token improves by $0.6\times$, $2.2\times$, $5.1\times$, and $10.4\times$ on Pixel 8Pro. This yields a favorable accuracy-latency Pareto frontier for on-device LLM inference in resource-constrained environments.
In contrast, GSM8K, a math-intensive benchmark requiring multi-step reasoning, exhibits the most polarized trade-off.  Smaller models yield high decoding efficiency but produce unusably low accuracy, while larger models offer acceptable accuracy only at the cost of prohibitively high latency. This sharp dichotomy underscores the difficulty in supporting GSM8K efficiently on-device without task-specific optimizations.
HellaSwag, which involves longer, narrative-style prompts and commonsense reasoning, falls between these two extremes. Notably, a clear ``most-balanced'' model, \texttt{SmolLM2-360M}, emerges near the knee of the Pareto frontier.
Finally, even within the same task category, task difficulty can significantly shape the trade-off. Compared to ARC-Easy, ARC-Challenge exhibits a flatter accuracy-latency frontier, where accuracy improves more slowly with increased latency, and smaller models rapidly encounter diminishing returns. 

\begin{graybox}
\underline{Implication 2:} \textit{These observations suggest that task- and difficulty-aware adaptive model selection might be crucial for achieving an optimal balance between performance and efficiency in on-device LLM deployments. Moreover, not all task categories are suitable for local processing on mobile and edge devices. Awareness of task type and difficulty can also inform decisions about when and how to offload computation to edge or cloud infrastructures, enabling effective device-server collaboration for LLM inference.}
\end{graybox}

Third, \textit{cross-suite differences hint at model architectural effects, beyond mere parameter scaling.} 
While scaling laws suggest that increasing model parameter size is the primary driver of LLM performance~\cite{kaplan2020scaling}, our results show that architectural design is equally critical, especially for compact models in on-device settings, in shaping the accuracy-latency trade-off.
For instance, \texttt{SmolLM2-360M} consistently matches or outperforms \texttt{Pythia-1B}, a model nearly three times larger, across all tasks. Notably, \texttt{SmolLM2} suite advances the Pareto frontier on benchmarks, such as ARC-Challenge and HellaSwag, achieving higher accuracy with lower decode latency.
This is primarily enabled by \texttt{SmolLM2}’s architecture and pretraining data curation,
both carefully optimized for resource-constrained, on-device inference \cite{allal2025smollm2,liu2024mobilellm}.

\begin{graybox}
\underline{Implication 3:} \textit{These observations suggest that model architecture, beyond parameter scaling, is a key factor shaping performance-efficiency trade-offs for compact on-device language models (i.e., sub-billion to billion-scale).
Targeted architectural design or search prior to pre-training may shift the Pareto frontier as effectively as scaling model size.
}
\end{graybox}

\textbf{Quantization impact on accuracy-latency trade-offs.} 
Model quantization is a widely adopted technique that reduces the precision of weights and activations in LLMs, thereby lowering memory footprint and computational cost. However, its practical impact on on-device LLMs remains underexplored, especially in terms of how quantization shifts the trade-off between accuracy and runtime latency. Prior studies has largely focused on theoretical metrics, such as FLOPs or parameters, while overlooking systematic empirical evaluation on this perspective.
To address this gap, we propose a new evaluation metric, the \textit{Harmonic Quantization score} ($\mathcal{HQ}$), and conduct a benchmarking study across the aforementioned model suites, tasks, and mobile platforms. Our goal is to systematically quantify the impact of quantization on both accuracy and runtime latency. We define the $\mathcal{HQ}$ for task $i$:
\begin{small}
\begin{equation}
\mathcal{HQ}_{i} = \left( \frac{1}{3} \left( \frac{1}{\mathcal{M}_{a,i}} + \frac{1}{\mathcal{M}^{\text{prefill}}_{l,i}} + \frac{1}{\mathcal{M}^{\text{decode}}_{l,i}} \right) \right)^{-1},
\label{eq:hes}
\end{equation}
\end{small}

\noindent
where $\mathcal{M}_{a,i} = \frac{A_{i}^{c}}{A_{i}}$, with $A_{i}^{c}$ and $A_{i}$ representing the normalized accuracy of the quantized and original models on task $i$, respectively. Similarly, $\mathcal{M}^{p}_{l,i} = \frac{L_{i,p}}{L_{i,p}^{c}}$, where $L_{i,p}^{c}$ and $L_{i,p}$ are the runtime latencies of the quantized and original model during inference phase $p\in\{\text{prefill}, \text{decode}\}$. The $\mathcal{HQ}$ rewards quantized models that achieve balanced accuracy and latency. Its harmonic mean formulation penalizes any severe degradation in individual components, thus discouraging improvements that come at the expense of others.

\begin{figure}[t]
\centerline
{\includegraphics[width=0.495\textwidth]{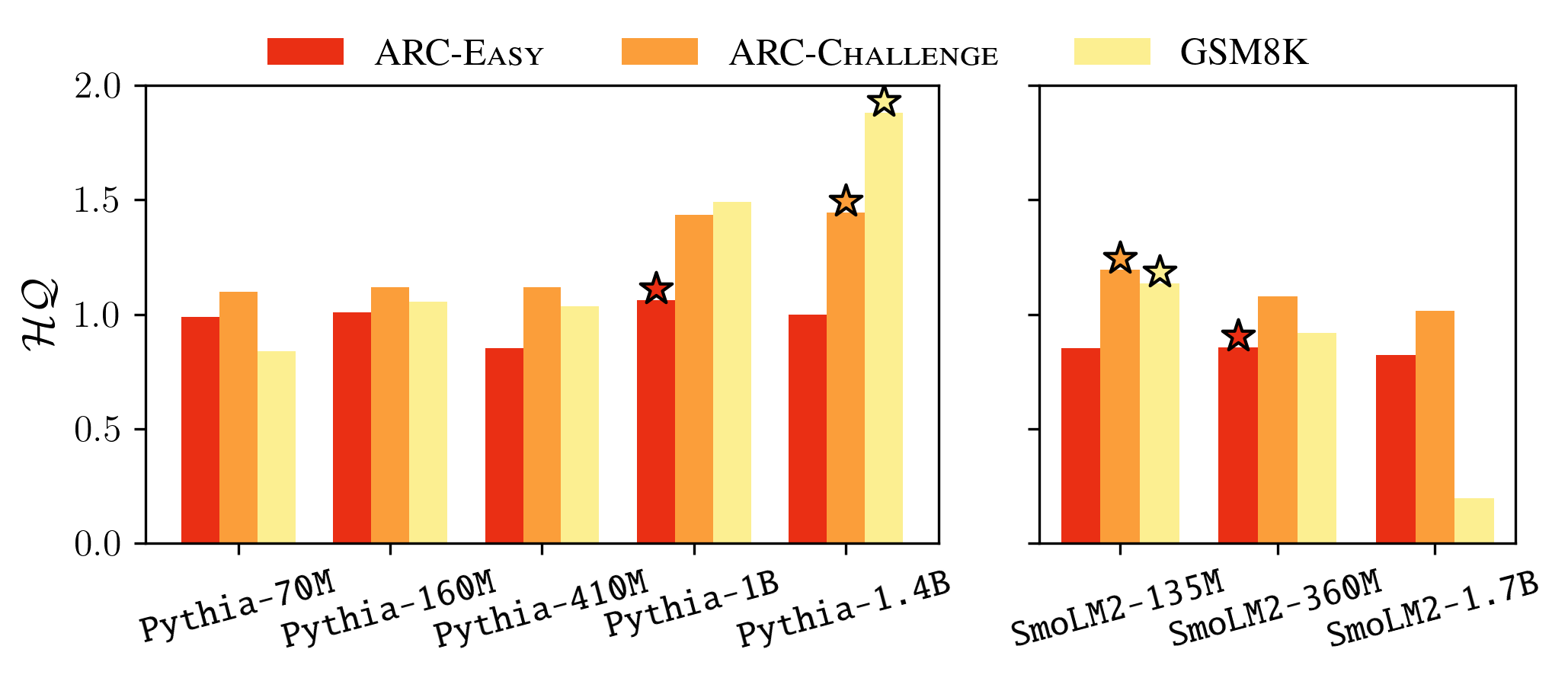}}
\caption{Task-wise Harmonic Quantization ($\mathcal{H}Q$) scores (higher is better) across model sizes for \texttt{Pythia} and \texttt{SmoLM2}. Stars mark the best-balanced model per suite. \texttt{Pythia} benefits from scaling up to 1.4B, while \texttt{SmoLM2} peaks at 135M-360M.}
\label{fig:quant}  
\end{figure}

We evaluate 4-bit quantized models from the \texttt{Pythia} and \texttt{SmoLM2} suites across multiple tasks. Fig. \ref{fig:quant} compares their $\mathcal{HQ}$ scores. First, \textit{quantization-induced efficiency gains vary across model architectures.} For \texttt{Pythia}, $\mathcal{HQ}$ increases nearly monotonically with model size, peaking at 1.4B for the more challenging tasks (ARC-Challenge and GSM8K), while saturating for ARC-Easy beyond 1B. 
In contrast, \texttt{SmoLM2} achieves its best accuracy-latency trade-off at smaller scales (135M-360M). 
Second, \textit{quantization benefits are task-dependent and appear correlated with task difficulty.} The optimal model size for quantization varies not only across architectures, but also across tasks within the same suite.
For \texttt{Pythia}, the average $\mathcal{HQ}$ rises from 0.982 on ARC-Easy to 1.243 on ARC-Challenge and 1.260 on GSM8K, indicating that 4-bit quantization yields limited benefit on simpler tasks, but delivers clear gains on demanding ones.
One anomalous case is \texttt{SmoLM2-1.7B} on GSM8K, whose $\mathcal{HQ}$ collapses due to near-zero accuracy reported by the \textit{lm-evaluation-harness}. We have evaluated three different 4-bit quantization schemes for this model, all yielding similarly poor accuracy. A plausible explanation is that aggressive 4-bit quantization introduces significant noise into numerically sensitive layers, such as those performing multi-step arithmetic.


\begin{graybox}
\underline{Implication 4:} \textit{These results underscore the value of our proposed $\mathcal{HQ}$ metric for benchmarking quantization techniques in on-device settings. Relying solely on accuracy or latency may hide critical efficiency regressions that impact on-device deployment. We consider evaluating diverse quantization techniques using $\mathcal{HQ}$ as our future work.
}

\noindent\underline{Implication 5:} \textit{
The presence of architecture- and task-specific sweet spots implies that quantization is not universally beneficial for on-device LLMs. The choice between quantized and full-precision models should jointly consider model size, task type, and hardware constraints.
}
\end{graybox}

\begin{figure*}[t]
\centering
\subfigure[Timeline view of kernel executions and data movements within a decode step.]
{\includegraphics[width=0.53\textwidth]{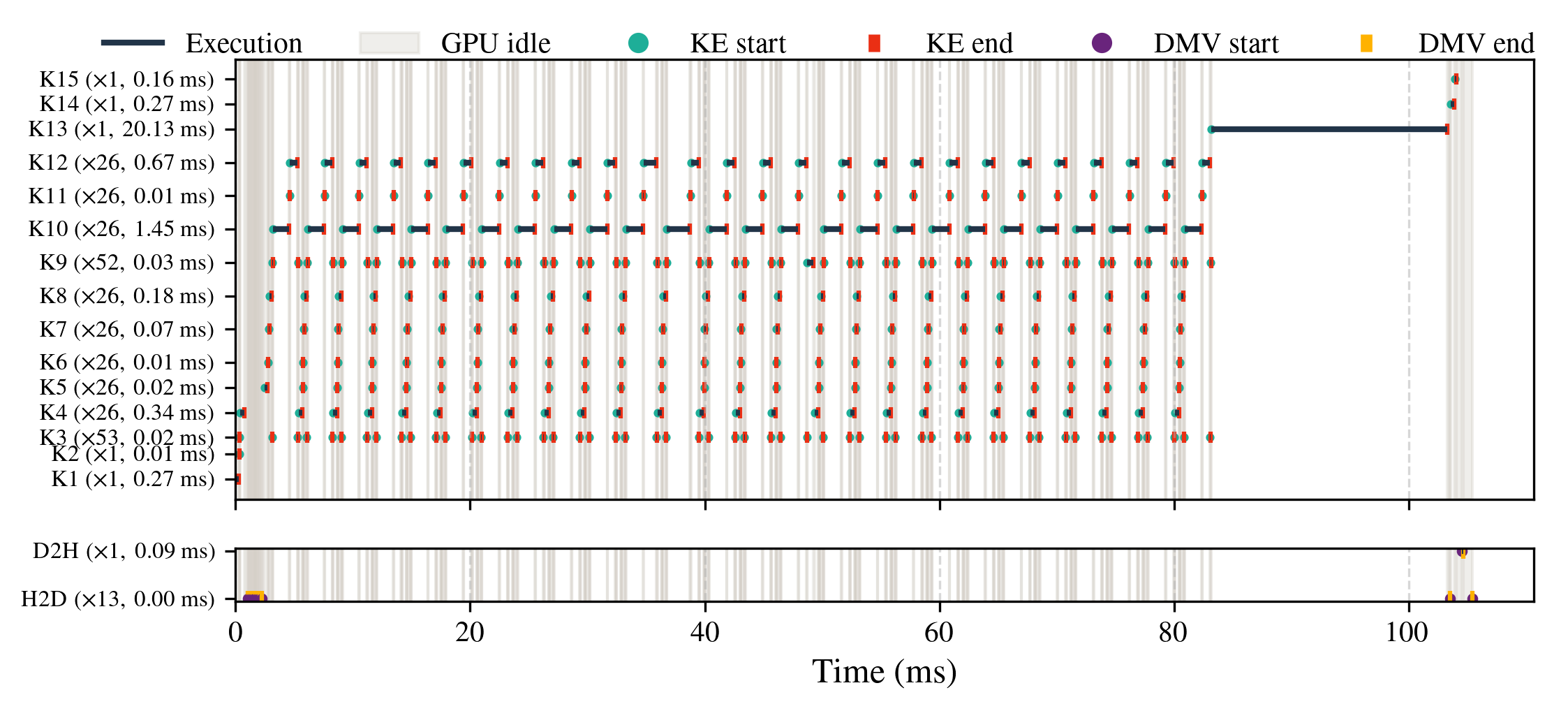}\label{fig:timeline}}
\subfigure[Breakdown of kernel execution time and GPU idle time.]
{\includegraphics[width=0.46\textwidth]{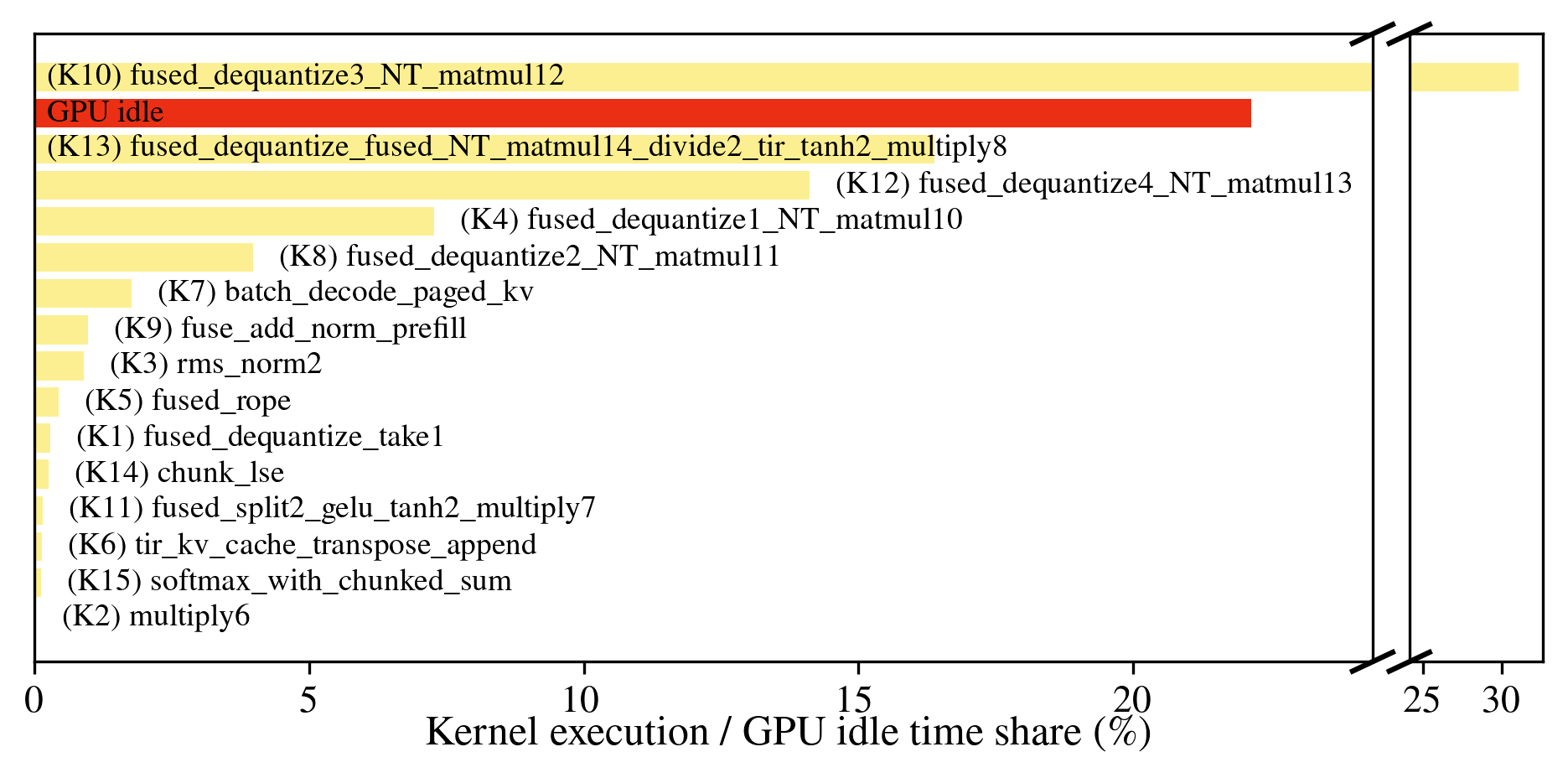}\label{fig:share}}
\caption{Visualization of kernel execution (KE), data movement (DMV), and GPU utilization in a single decode step. (a) Each kernel (y-axis) is labeled by index, invocation count, and average runtime; gray regions indicate GPU idle periods. (b) Kernel names are annotated with indices; red bar shows the GPU idle proportion during decoding.}
\label{fig:kernel}  
\end{figure*}

\begin{figure}[t]
\centerline
{\includegraphics[width=0.48\textwidth]{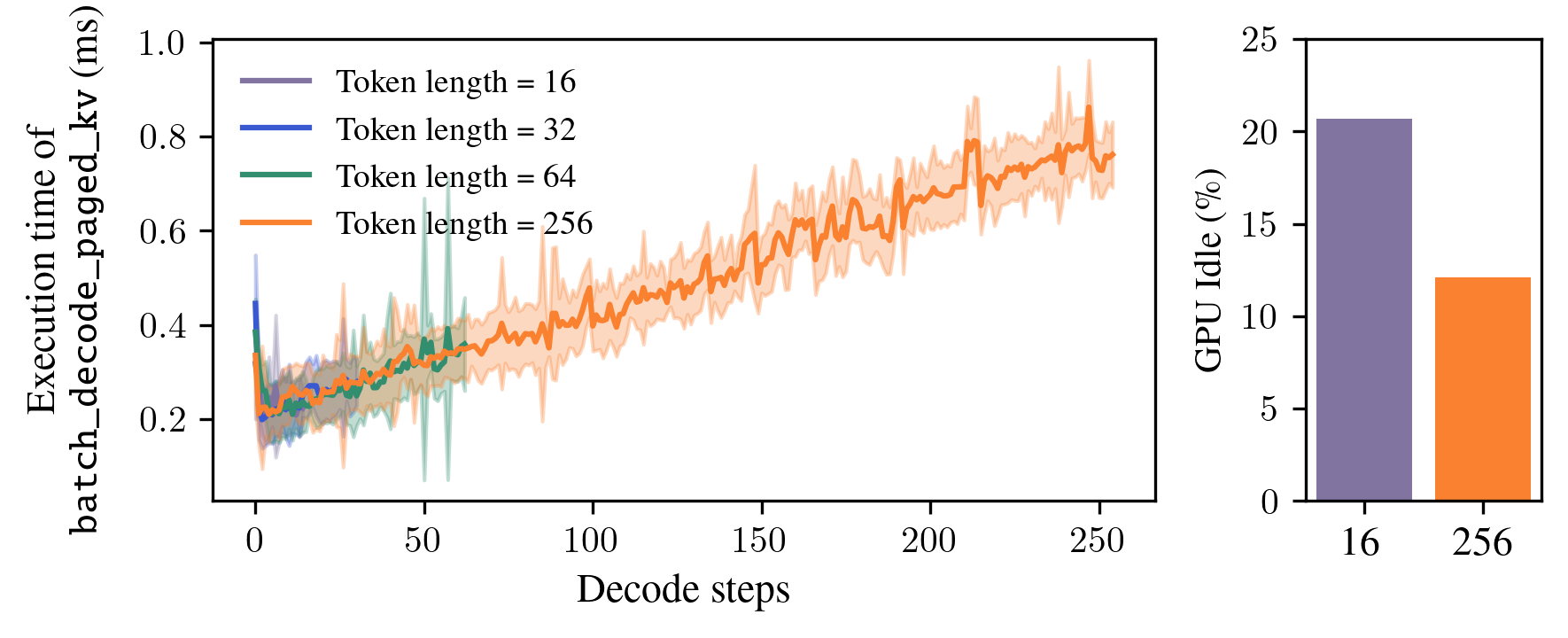}}
\caption{Paged-attention kernel latency grows with decode steps, while GPU idle time drops with longer outputs. Each step generates one token.} 
\label{fig:pagekv}  
\end{figure}

\subsection{Kernel-oriented Empirical Study}
\label{ssc:kernel}
We then employ \textsc{lm-Meter} for a kernel-level analysis of on-device LLM execution. Since kernels serve as the fundamental scheduling units and embed numerous framework-level optimizations, profiling them provides deep visibility into the true performance landscape and enables characterization of fine-grained runtime behavior. Our empirical study is guided by the following research questions: (i) Which kernels dominate runtime latency? (ii) Do these dominant kernels remain consistent bottlenecks throughout the decode sequence, or do new ones emerge as the KV cache grows? (iii) How much of the decode timeline is spent in GPU-idle states? (iv) Can the observed kernel behaviors be leveraged to predict per-step decode latency for scheduling and runtime adaptation?
Due to the page limit, we present representative results from running a 4-bit quantized \texttt{Gemma-2-2B-it} on a Pixel 8 Pro. These insights expose several system-level inefficiencies and optimization opportunities, many of which generalize to other models and platforms.


\textbf{Kernel runtime performance.} Fig.~\ref{fig:timeline} depicts fine-grained start/end timestamps of GPU kernel executions within a single decode step, illustrating how kernels are sequenced and interleaved.
Notably, frequent gaps between successive kernel launches (gray regions) lead to over 21\% GPU idle time, making it the second-largest contributor to total latency. These idle periods are primarily attributed to host-side data preparation delays and I/O stalls.
Additionally, multiple memory-bound short-duration kernels, including \texttt{fused\_rope} (K5), \texttt{tir\_kv\_cache\_transpose\_append} (K6), \texttt{fuse\_add\_norm\_prefill} (K9), and \texttt{rms\_norms} (K3), are invoked frequently yet contribute less than $10\%$ of total execution time. 
Their high invocation frequency, however, increases kernel launch overhead and widens idle gaps, exposing inefficiencies in kernel scheduling and dispatch. Moreover, three fused GEMM kernels (K10, K12, K13) dominate GPU execution, contributing over 60\% of total latency. This confirms that MLP projections are primary compute bottleneck during decoding, especially for short sequences, surpassing attention-related kernels (K4, K8).


\begin{graybox}
\underline{Implication 6:} \textit{Frequent GPU idle periods reveal opportunities to enhance system performance and efficiency. Runtime systems can downscale GPU frequency during idle windows to save power or overlap lightweight computations to boost utilization. Moreover, the abundance of short, fragmented kernels underscores the need for improved kernel fusion strategies tailored to on-device LLMs.}
\end{graybox}

\textbf{Paged-attention kernel latency.} Fig. \ref{fig:pagekv} depicts the average latency of the \texttt{batch\_decode\_paged\_kv} kernel over decode steps for four target output token sequence lengths. This kernel is the core operator for paged attention~\cite{kwon2023efficient}. We observe that its latency grows nearly linearly with decode steps, as each invocation must scan an increasingly large KV cache. The growth is steeper for longer outputs, reaching up to 0.9 ms per token by step 250 in the 256-token case. 
The right panel compares total GPU idle time for 16- and 256-token generations. Interesting, generating longer sequences reduces idle time from 21\% to 12\%. These results indicate that long-sequence generation may be ill-suited for resource-constrained devices due to rising per-step latency.
Conversely, short-sequence generations leave significant headroom for optimizing GPU utilization efficiency.

\begin{graybox}
\underline{Implication 7:} \textit{Among all decode-phase kernels, we observe that \texttt{batch\_decode\_paged\_kv} is the only one whose latency scales with decode steps. Modeling this trend enables lightweight online prediction of per-step latency, supporting efficient scheduling and improved responsiveness in on-device LLM systems.}
\end{graybox}


\subsection{Limitations}

This work offers a first step in characterizing on-device LLM runtime behavior but is limited to a subset of mobile SoCs. Other edge platforms (Jetsons and Intel NPUs) may exhibit different bottlenecks and optimization opportunities.
Our compression analysis focuses on post-training 4-bit quantization, a popular yet narrow slice of the broader compression landscape. Techniques like structured pruning, activation sparsity, and hybrid-precision quantization may yield different trade-offs depending on hardware and workload.


\section{Related Work}
\label{sc:related}


{\sloppy
Recent advances in model compression and compiler optimizations have made on-device LLM deployment increasingly feasible. Techniques such as quantization~\cite{shao2023omniquant, xiao2023smoothquant, frantar2022gptq, lin2024awq}, knowledge distillation~\cite{gu2023minillm}, and sparsification~\cite{frantar2023sparsegpt, sun2023simple, ma2023llm} reduce model size and computation while preserving accuracy.
ML compilers like MLC~\cite{mlc}, LLMFarm~\cite{llmfarmer}, TensorRT-LLM~\cite{TensorRT}, and llama.cpp~\cite{llamacpp} unify LLM inference across heterogeneous edge platforms, via kernel fusion, quantization-aware compilation, and hardware-specific tuning.
However, existing evaluations of on-device LLM systems largely focus on end-to-end latency or task-level accuracy, offering limited insight into phase- or kernel-level bottlenecks. Our work complements this direction by introducing a lightweight, online profiler that enables fine-grained decomposition of inference latency, providing deeper visibility into runtime behavior and guiding system-level optimization strategies.
\par
}

\section{Conclusion}

In this paper, we present \textsc{LM-Meter}, a lightweight online profiler that provides accurate phase- and kernel-level latency measurements. Using \textsc{lm-Meter}, we conduct empirical studies and uncover actionable insights: prefill is the dominant on-device bottleneck, accuracy-latency trade-offs are highly task- and architecture-dependent, and our Harmonic Quantization score offers a holistic view of quantization impact. Kernel-level profiling also reveals substantial GPU idle time and highlights opportunities for optimization.





\begin{acks}
We thank the reviewers and our shepherd for their insight-
ful comments. This work was supported by funds from Toyota Motor North America.
\end{acks}

\bibliographystyle{unsrt}
\bibliography{references}

\end{document}